\documentclass[letterpaper, 10 pt, conference]{format/ieeeconf} % letterpaper or a4paper
\IEEEoverridecommandlockouts       
\usepackage[normalem]{ulem}	                        % underlining!
\usepackage[table,usenames,dvipsnames]{xcolor}      % color
\usepackage{extarrows}                              % http://ctan.org/pkg
\let\labelindent\relax
\usepackage{enumitem}                               % [inline]

\usepackage{amsmath,amsthm,amssymb,amsfonts,dsfont} % math
\usepackage{cite}
\makeatletter
\newif\if@restonecol
\newif\foralgo@restonecol
\makeatother
\usepackage{graphicx}
\usepackage{subfigure}
\usepackage{tabularx}
\usepackage{booktabs}
\usepackage{multirow,multicol, array}
\usepackage[font={small}]{caption} 

\usepackage{algorithm,algorithmic}
\let\appendices\relax
\usepackage[titletoc,title]{appendix}
\usepackage[breaklinks=true, colorlinks, bookmarks=true, citecolor=Black, urlcolor=Magenta,linkcolor=Black]{hyperref}
\usepackage{cleveref}

\usepackage{setspace}
\theoremstyle{definition}

\newtheorem{definition}{Definition}
\newtheorem{assumption}{Assumption}
\newtheorem*{assumption*}{Assumption}

\newtheorem*{remark*}{Remark}
\newtheorem*{problem*}{Problem}

\title{\LARGE \bf
Skill Transfer and Discovery for Sim-to-Real Learning: \\
A Representation-Based Viewpoint 
}
\author{Haitong Ma, Zhaolin Ren, Bo Dai, Na Li
\thanks{Haitong Ma, Zhaolin Ren, Na Li are with School of Engineering and Applied Sciences, Harvard University. Bo Dai is with School of Computational Science and Engineering, Georgia Institute of Technology. Email: \texttt{\{haitongma, zhaolinren\}@g.harvard.edu, bodai@google.com, nali@seas.harvard.edu}. The work is supported under NSF AI institute: 2112085, NSF CNS: 2003111, NSF ECCS: 2328241}
\thanks{The Github link for codes, videos, and full report is available at \href{https://congharvard.github.io/steady-sim-to-real/}{https://congharvard.github.io/steady-sim-to-real/}}
}

\begin{document}
\maketitle
\begin{abstract}
    We study sim-to-real skill transfer and discovery in the context of robotics control 
    % \haitong{Some representation-based transfer learning papers focus on the vision model and deep learning, trying to separate our paper from those}
    using representation learning. We draw inspiration from spectral decomposition of Markov decision processes. The spectral decomposition brings about representation that can
    % the analysis of linear Markov decision processes, where representations coming from the spectral decomposition of transition dynamics can be regarded as skill sets. 
    %The representations can 
    linearly represent the state-action value function induced by any policies, thus can be regarded as skills. The skill representations are transferable across arbitrary tasks with the same transition dynamics. 
    Moreover, to handle the sim-to-real gap in the dynamics, we propose a skill discovery algorithm that learns new skills caused by the sim-to-real gap from real-world data. We promote the discovery of new skills by enforcing orthogonal constraints between the skills to learn and the skills from simulators, and then synthesize the policy using the enlarged skill sets. We demonstrate our methodology by transferring quadrotor controllers from simulators to Crazyflie 2.1 quadrotors. We show that we can learn the skill representations from a single simulator task and transfer these to multiple different real-world tasks including hovering, taking off, landing and trajectory tracking. Our skill discovery approach helps narrow the sim-to-real gap and improve the real-world controller performance by up to 30.2\%.
\end{abstract}

\setlength{\abovedisplayskip}{5pt}
\setlength{\abovedisplayshortskip}{5pt}
\setlength{\belowdisplayskip}{5pt}
\setlength{\belowdisplayshortskip}{5pt}

\newcommand{\simulatorv}{\widehat{V}}
\newcommand{\simulatorp}{{P}^\circ}
\newcommand{\simulator}[1]{{#1}^{\circ}}
\newcommand{\innerprod}[2]{\left\langle{#1},{#2}\right\rangle}
\newcommand{\augphi}{\phi}
\newcommand{\augmu}{\mu}
\newcommand{\wholephi}{\bar\phi}
\newcommand{\wholemu}{\bar\mu}
\newcommand{\orthdim}{s}
\renewcommand{\algorithmicrequire}{\textbf{Input:}}
\renewcommand{\algorithmicensure}{\textbf{Output:}}
\section{Introduction}
Reinforcement learning (RL) has demonstrated superior performance in many robotic simulators \cite{silver_mastering_2017,haarnoja_soft_2018,wurman_outracing_2022}. However, transferring the controllers learned in simulators to real robots has long been a very challenging question in the RL community. One difficulty of sim-to-real transfer is that the learned policies are highly specific to the dynamics and tasks in the simulators, making them difficult to generalize to many real-world tasks, in which sim-to-real gaps exist. In the existing practices of sim-to-real transfer % typically only transfer the controllers
methods~\cite{andrychowicz_learning_2020,kaufmann_champion-level_2023,molchanov_sim--multi-real_2019,gronauer_comparing_2023}, the controllers learned in the simulator are usually difficult to generalize across various tasks and dynamic environments.

% The most commonly used sim-to-real transfer technique is domain randomization \cite{tobin_domain_2017,molchanov_sim--multi-real_2019,andrychowicz_learning_2020}, where the physical parameters of the simulator dynamics are randomly perturbed. However, the perturbation rules rely on human knowledge and 

% The most commonly used method to tackle the differences in dynamics should be domain randomization \cite{tobin_domain_2017}. Domain randomization in robotics usually means that the simulator randomly generates physical parameters in the dynamics model, rather than using only one specific dynamics. For example, \citet{andrychowicz_learning_2020} randomize the mass, friction and damping coefficients, 

Recent studies in the theoretical RL community on the spectral decomposition of Markov decision processes (MDPs) \cite{jin_provably_2019,yang_reinforcement_2020, agarwal_flambe_2020,uehara_representation_2022,ren_spectral_2023} reveal the idea of task-independent representations for RL. The results of spectral decomposition is the spectral functions of the transition dynamics in MDPs. The spectral functions can \emph{linearly}  represent the state-action value function, i.e., the $Q$-function, \emph{induced by any policy}. Therefore, we say the spectral functions are task-independent representation of \textit{skills}, because the spectral functions include information needed to accomplish any tasks.
These task-independent skill representations are shared across arbitrary tasks, thus are reusable and transferable. Meanwhile, the representation can also be used to synthesize policies. 
% If we know the representations, fitting the Q function for a specific task only needs solving a linear regression. 
Given the representation-based skill sets and a specific task, we can perform sample-efficient planning upon the skill sets to synthesize the optimal policy. When the representations are unknown, sample-efficient representation learning methods have been proposed including maximum likelihood estimation \cite{uehara_representation_2022}, contrastive learning~\cite{zhang_making_2022}, spectral conditional density estimation \cite{ren_spectral_2023}, or variational inference \cite{ren_latent_2023}. 

Nevertheless, these representation-based skill learning are still designed for specific transition dynamics. When it applies to sim-to-real transfer, 
the sim-to-real gap, which will induce new skills different from the simulator skill sets, has not been investigated yet. 
Learning the sim-to-real gap from real-world data, also called residual dynamics learning \cite{saveriano_data-efficient_2017,kaufmann_champion-level_2023,johannink_residual_2019,shi_neural_2019}, naturally aligns with our representation learning viewpoint. However, naively learning the representations of residual dynamics might lead us to relearn redundant skills that are linearly dependent with the existing simulator skill sets. Therefore, we need additional incentives to \emph{discover} new skills that enable us to bridge the sim-to-real gap.

To further leverage the transferability of the representation-based skill sets and discover new skills induced by the sim-to-real gap, we proposed the \textbf{S}kill \textbf{T}ransf\textbf{E}r \textbf{A}nd \textbf{D}iscover\textbf{Y} \textbf{(STEADY)} for sim-to-real representation learning algorithm. We show that recent theoretical representation learning algorithms for spectral decomposition of MDPs, such as \cite{ren_spectral_2023,ren_stochastic_2023}, can apply to learning transferable representations of real-world robots. Moreover, we handle the sim-to-real gap by augmenting distinct representation-based skills learned from the sim-to-real gap, which we refer to as \emph{skill discovery}. During the learning process, orthogonal constraints between the newly discovered skill sets and the simulator skill sets are enforced to fill the sim-to-real gap.
% the skills learned from the sim-to-real gap are different from the skills the simulator already had. 
In this way, we ensure that the skills necessary for the real robots are also included in our augmented skill sets, upon which the planning can be handled in a more complete space efficiently. 
% {\color{blue} [I think ``augmented'' skill sets usually means ``enlarged'' skill sets (i.e. simulator + new discovered skills), so I modified this paragraph to reflect that. Previously, augmented skill sets seemed to refer to just the new skill sets, which may be confusing]}
Meanwhile, to ensure the policy transferring smoothly from the simulator to the real-world, we also designed a mechanism to characterize the policy shift in our algorithm. 

We demonstrate our proposed algorithm by transferring the learned quadrotor control policies to Crazyflie 2.1 quadrotors. First, we learn the representations  and train a tracking controller on the simulator. Then, we collect task-specific data, including hovering, taking-off, landing, and trajectory tracking in the real world for skill transfer and discovery. The results show that our representation-based skills are easily transferred to the real world and generalizable to different tasks, and that our method has improved the real-world tracking performance by up to 30.2\%.

\section{Related works}

\subsection{Sim-to-Real Transfer}
Sim-to-real transfer aims to transfer knowledge from simulator to real-world robots and overcome the sim-to-real gap. We focus on sim-to-real gaps in the dynamics, which might arise from measurement noises, defective actuators, inexact physical parameters, contact or fluid effects, response delays, etc. Representative sim-to-real transfer techniques include domain randomization and residual dynamics learning.
\subsubsection{Domain randomization}
 Domain randomization in RL refers to randomly perturbing the physical parameters of the simulators \cite{tobin_domain_2017,molchanov_sim--multi-real_2019,andrychowicz_learning_2020,kaufmann_champion-level_2023,chebotar_closing_2019,peng_sim--real_2018}. Then the RL agents aim to learn a policy performing well under a distribution of transition dynamics. The trained policies are directly applied to the real world and no knowledge will be learned from real-world data. These methods are convenient since simulator data are usually cheap to sample. However, domain randomization can only fix the sim-to-real gaps that can be modeled, such as inexact physical parameters. It cannot handle other sim-to-real gaps like aerodynamics, contact effects, response delays, etc.

\subsubsection{Residual dynamics learning} Learning the sim-to-real gap, or the residual dynamics, appears in many recent sim-to-real transfer studies \cite{levine_learning_2014,saveriano_data-efficient_2017,kaufmann_champion-level_2023,shi_neural_2019,shi_neural-swarm_2020,johannink_residual_2019}. Unlike the domain randomization, during the real-world implementation stage, real-world data are collected to learn the sim-to-real gap and improve the policies. Existing practices have considered different models, including Gaussian mixture model~\cite{levine_learning_2014}, Gaussian process \cite{saveriano_data-efficient_2017,fisac_general_2019}, $k$-nearest neighbors \cite{kaufmann_champion-level_2023}, or deep neural networks \cite{shi_neural_2019, shi_neural-swarm_2020,johannink_residual_2019}. Then the learned sim-to-real gap is integrated with the prior simulator knowledge as external disturbance \cite{shi_neural_2019,shi_neural-swarm_2020,fisac_general_2019} or additive controllers \cite{johannink_residual_2019}.

We also follow the residual dynamics learning idea to fill the sim-to-real gap, but through a novel representation view. 
We use the knowledge of the simulator skill sets in the residual dynamics learning process in the way that enforces orthogonal constraints between simulator skill sets and the new skill sets, expanding skill sets while preventing redundancy of learning skills that are already in the simulator skill sets.

\subsection{Representation-Based Knowledge Transfer} 
Representation-based knowledge transfer is commonly seen for computer vision models and visual-input control tasks\cite{du_auto-tuned_2021,tanwani_dirl_2021}.  For dynamics control tasks, \cite{devin_learning_2017} decomposed the policy networks into task-specific and robot-specific modules and show that the modules are transferable. \cite{helwa_multi-robot_2017} proposed transfer learning by matching the transfer functions, where transfer functions can be regarded as another type of representation. However, the algorithms are only limited to single-input-single-output dynamical systems. Our approach applies to general nonlinear dynamical systems and the transferability is justified rigorously by theoretical analysis. For theoretical representation learning, \cite{agarwal_provable_2023} has proved that there are provable benefits on sampling complexity for transferring the representations. However, no experimental results on simulators or in real world has been reported.

\subsection{Representation Learning via Spectral Decomposition}
% The sample complexity of MDP planning algorithms (like value iteration or policy iteration) scales with the sizes of the state and action spaces. Therefore, there exists no sample-efficient planning algorithm for general MDPs with continuous state and action spaces.
The spectral structrue in the dynamics of MDP paves the way for sample-efficient planning on MDPs with continuous state and action spaces \cite{jin_provably_2019, agarwal_flambe_2020}. However, the analysis relies on the known representation, i.e., the spectral decompositions of transition dynamics. But in practice, the representations are intractable for general dynamics. Then \cite{agarwal_flambe_2020,uehara_representation_2022} proposed conceptual algorithms upon some computational oracle that simultaneously learns the representations and solves the optimal policy. Later \cite{ren_spectral_2023,ren_latent_2023} further push the representation learning practical, using techniques like spectral conditional density estimation and variational learning, which provides us with practical algorithms for real-world robotic systems.

\section{Preliminaries}
\subsection{Notations and Sim-to-Real Problem Setting}
Markov Decision Processes (MDPs) are a standard sequential decision-making model for RL, and can be described as a tuple $\mathcal{M}=(\mathcal{S}, \mathcal{A}, r, P, \rho, \gamma)$, where $\mathcal{S}$ is the state space, $\mathcal{A}$ is the action space, $r: \mathcal{S} \times \mathcal{A} \to \mathbb R$ is the reward function, $P: \mathcal{S} \times \mathcal{A} \rightarrow \Delta(\mathcal{S})$ is the transition operator with $\Delta(\mathcal{S})$ as the family of distributions over $\mathcal{S}, \rho \in \Delta(\mathcal{S})$ is the initial distribution and $\gamma \in(0,1)$ is the discount factor. The goal of RL is to find a policy $\pi: \mathcal{S} \rightarrow \Delta(\mathcal{A})$ that maximizes the infinite-horizon cumulative discounted reward $$\mathbb{E}_{s_0 \sim \rho, \pi}\left[\sum_{i=0}^{\infty} \gamma^i r\left(s_i, a_i\right) \mid s_0\right]$$ by interacting with the MDP.  
The value function under transition dynamics $P$ and policy $\pi$ is defined as $V_{P}^\pi(s)=\mathbb{E}_\pi\left[\sum_{i=0}^{\infty} \gamma^i r\left(s_i, a_i\right) \mid s_0=s\right]$, and the state-action value function under transition dynamics $P$ is $Q_{P}^\pi(s, a)=\mathbb{E}_\pi\left[\sum_{i=0}^{\infty} \gamma^i r\left(s_i, a_i\right) \mid s_0=s, a_0=a\right]$. 
When doing off-policy learning we slightly abuse the notation of $\mathcal{B}$ to denote any data distribution sampled from the off-policy data set of replay buffer. 

The sim-to-real problem indicates that we have a simulator of the real-world MDP $\mathcal{M}$, which is also an MDP $\simulator{\mathcal{M}}=(\mathcal{S}, \mathcal{A}, \simulator r, \simulatorp, \simulator\rho, \gamma)$. Notations with superscript ${}^\circ$ means they are related to the simulator. The two MDPs might differ in the transition dynamics, $\simulator{P}$ and ${P}$, initial distributions $\simulator\rho,\rho$ and rewards $\simulator r,r$. The simulator is cheap to query and the real world is expensive. Therefore, we split the learning procedure into the simulator stage and the real-world transfer stage. In the simulator stage, the agent interacts with $\simulator{\mathcal{M}}$ for sufficiently many transitions to obtain knowledge from the simulators. Then in the real-world stage, the agent transfers knowledge learned in the simulator to solve the optimal policy of $\mathcal{M}$, while only collecting a limited number of transitions by interacting with $\mathcal{M}$.

\subsection{Spectral Decomposition and Skills in Markov Decision Processes}
% We additionally define the state visitation distribution induced by a policy $\pi$ as $d_P^\pi(s)=(1-$ $\gamma) \mathbb{E}_{s_0 \sim \rho, \pi} \mathbb{E}\left[\sum_{t=0}^{\infty} \gamma^t \mathbf{1}\left(s_t=s\right) \mid s_0\right]$, where $\mathbf{1}(\cdot)$ is the indicator function.
% When $|\mathcal{S}|$ and $|\mathcal{A}|$ are finite, there exist sample-efficient algorithms that find the optimal policy by maintaining an estimate of $P$ or $Q_{P, r}^\pi$ \cite{jin_is_2018, azar_minimax_2017}. However, such methods cannot be scaled up when $|\mathcal{S}|$ and $|\mathcal{A}|$ are extremely large or infinite. 
% Function approximation is needed to exploit the structure of the MDP when $|\mathcal{S}|$ and $|\mathcal{A}|$ are extremely large or infinite. 
% Recently, sample-efficient algorithms for MDPs with large or infinite state and action spaces have been proposed with assumptions on the MDP structure. 

% is one of the most prominent structures that allows for simple yet effective function approximation in MDPs, which is based on the following spectral
The formal definition of spectral decomposition of MDPs refers to the following structures on the transition dynamics and rewards,

\begin{definition}[Spectral decomposition of MDPs,~\cite{jin_provably_2019,agarwal_flambe_2020}]\label{def:linearmdp} The spectral decomposition of an MDP $\simulator{\mathcal{M}}$ with transition dynamics $\simulator{P}\left(s^{\prime} \mid s, a\right)$ means there exists representations $\simulator{\phi}: \mathcal{S} \times \mathcal{A} \rightarrow \mathbb{R}^d$ and $\simulator{\mu}: \mathcal{S} \rightarrow \mathbb{R}^d$ such that
$$
\simulator{P}\left(s^{\prime} \mid s, a\right)=\left\langle\simulator{\phi}(s, a), \simulator{\mu}\left(s^{\prime}\right)\right\rangle, \quad r(s, a)=\left\langle\simulator{\phi}(s, a), \theta_r\right\rangle
$$
where $\theta_r \in \mathbb{R}^d$ and $\langle\cdot,\cdot\rangle$ denotes the vector inner product.
\end{definition}
The spectral decomposition enables that the representation $\phi$ can linearly represent the state-action value function $Q^\pi_{\simulatorp}$ \emph{for any policy} $\pi$,
\begin{equation}
    Q_{\simulator{P}}^\pi(s, a)=\left\langle\simulator{\phi}(s,a),w^\pi\right\rangle \label{eq:linear_q}
\end{equation}
where $w^\pi = \theta_r+\gamma \int_\mathcal{S}V_{\simulatorp}^\pi\left(s^{\prime}\right) \mu\left(s^{\prime}\right) d s^{\prime}$. The linear structure can be obtained by the recursive relationship between the $Q_{\simulator{P}}^\pi$ and $V_{\simulator{P}}^\pi$,
$$
\begin{aligned}
     Q_{\simulator{P}}^\pi(s, a) = & r(s, a) + \gamma\int_{\mathcal{S}}\simulatorp(s'|s, a)V_{\simulatorp}^\pi(s')ds'\\
     = & \innerprod{\simulator{\phi}(s, a)}{\theta_r} + \gamma\int_{\mathcal{S}}\innerprod{\simulator{\phi}(s, a)}{\simulator{\mu}(s')}V_{\simulatorp}^\pi(s')ds'\\
     = & \innerprod{\simulator{\phi}(s, a)}{\underbrace{\theta_r+\gamma \int_\mathcal{S}V_{\simulatorp}^\pi\left(s^{\prime}\right) \simulator{\mu}\left(s^{\prime}\right) d s^{\prime}}_{w^\pi}}.
\end{aligned}
$$
\vspace{-5pt}

\noindent\textbf{Representation-based skills.} Note that the linear structure of $Q$-function holds for any policies under dynamics $\simulatorp$. We argue that $\simulator\phi$ can be regarded as the skill sets under dynamics $\simulatorp$, since it includes the information of constructing arbitrary policy. Therefore, we interpret the representation $\simulator\phi$ as the task-independent \emph{skill sets}, which includes the information of all the skills needed for the model $\simulatorp$. 
% Take our drones as example, the same $\simulator\phi$ could be used for multiple tasks like hovering and tracking any types of position or velocity trajectories\footnote{For taking off and landing, since the drones have the ground effect, the dynamics is different. We need to use the skill discovery to handle it, which will be described in the following sections.}.  
Formally, given a $Q$-function, it induces the max-entropy policy as 
% \begin{equation}
\begin{align}
    \pi_Q(a \mid s):= & \frac{\exp \left(\frac{Q(s, a)}{\tau}\right)}{\sum_{a \in \mathcal{A}} \exp \left(\frac{Q(s, a)}{\tau}\right)}\\
    =&\underset{\pi(\cdot \mid s) \in \Delta(\mathcal{A})}{\arg \max } \mathbb{E}_\pi[Q(s, a)]+\tau H(\pi)\label{eq:policy_phi}
\end{align}
% \end{equation}
where $\pi_Q$ is the greedy max-entropy policy given a $Q$-function, $H(\pi):=\sum_{a \in \mathcal{A}} \pi(a \mid s) \log \pi(a \mid s)$. Therefore, if we know the skill sets $\simulator\phi$, we can construct the max-entropy policies $\pi(a\mid s)\propto \operatorname{exp}(w^\top\simulator\phi(s, a))$ from skills $\phi^\circ(s, a)$. 

\vspace{5pt}
\noindent\textbf{Practical Implementations.} For practical implementations, we can parameterize the $Q$-function as $Q(s,a;w) = w^\top\simulator\phi(s, a)$. Then, the policy evaluation can be conducted by minimizing the temporal-difference error w.r.t. the parameter $w$, i.e., 
\begin{equation}
    \min_w \mathbb E_{(s,a,r,s')\sim\mathcal{B},a'\sim\pi(\cdot\mid s')}[(r + \gamma \overline Q(s',a') - w^\top\simulator\phi(s, a))^2] \label{eq:pev}
\end{equation}
where $\mathcal{B}$ is the data distribution in the replay buffer, and $\overline{Q}$ is the target $Q$-function commonly used in the target network trick. We emphasize that comparing to the deep Q-learning~\cite{mnih_playing_2013}, the policy evaluation optimization is in the linear space spanned by the learned skill sets, therefore, is more stable. 
The policy improvement is the same as in other off-policy algorithms that optimize \eqref{eq:policy_phi} by policy gradient. Practical implementations like soft actor-critic uses the reparameterization trick to simplify the calculations of \eqref{eq:policy_phi} \cite{haarnoja_soft_2018}. We will discuss our policy parameterization in the experimental setup in Section \ref{sec:exp}.
% \begin{remark}[$\simulator{\phi}$ as function bases.]\label{remark:space}
%     Intuitively, $\phi(s, a)$ can be regarded as $d$ functions $\phi_1,\phi_2,\dots,\phi_d$ serving as a set of bases of the Q-function $Q^\pi_{\simulatorp}(s, a)$. The linear representation means for any policy $\pi$, the value functions $Q^\pi_{\simulatorp}$ lies in the linear span of these $d$ fucntion bases $\operatorname{span}\{\phi_1,\phi_2,\dots,\phi_d\}$.
% \end{remark}

% If we know the $\simulator{\phi}(s, a)$, the practical implementation with low-rank representations only modify the critic parameterization to be  
% $
% Q(s, a;\theta) = \langle\simulator{\phi}(s, a), \theta \rangle
% $ with parameters $\theta$. Therefore, we can optimize the critic by optimizing the least-square loss, for example,
% \begin{equation}
% 	\mathbb E_{(s, a, r, s')\sim \mathcal D, a'\sim\pi(\cdot\mid s')}\left[\left(r + \bar Q(s', a') - Q(s, a;\theta)\right) ^2\right]
% \end{equation}
% where $\bar Q$ is the target Q-function.
% \bo{add some discussion why the representation can be viewed as skills.}

\subsection{Skill Learning by Spectral Conditional Density Estimation}
If the representation $\simulator{\phi}, \simulator{\mu}$ is unknown, we need to learn these representations from data. The objective is to make $\langle\simulator{\phi}(s, a), \simulator{\mu}(s')\rangle$ as close as $\simulator{P}(s'\mid s, a)$ as possible. There are multiple statistical learning methods to solve the problem~\cite{zhang_making_2022,ren_latent_2023,ren_spectral_2023}. 
% like noise contrastive learning used in \cite{zhang_making_2022} or spectral conditional density estimation \cite{ren_spectral_2023}. 
We follow the spectral conditional density estimation method, which optimizes the following loss function,
\begin{equation}
	\min_{\simulator{\phi}, \simulator{\mu}}\mathbb E_{(s,a)\sim\mathcal{B},s'\sim \simulator{P}(\cdot\mid s,a)}\left[\left\|\simulatorp(s'\mid s, a) \!-\! \innerprod{\simulator{\phi}(s, a)}{\simulator{\mu}(s')}\right\|_2^2\right]\label{eq:least_square_simp}
\end{equation}
where $\mathcal{B}$ denotes the offline dataset distribution.
However, \eqref{eq:least_square_simp} is usually intractable since we usually do not know the exact value of $\simulatorp(s'\mid s, a)$. This necessitates the use of sampling-based algorithms. For example,  \cite{ren_spectral_2023} uses a surrogate loss function that is equivalent with \eqref{eq:least_square_simp}:
\begin{equation}
\begin{aligned}
 L_{\text{feat}}(\simulator\phi, \simulator\mu):=&C-2 \mathbb{E}_{(s, a) \sim \mathcal{B}, s^{\prime} \sim P\left(s^{\prime} \mid s, a\right)}\left[\phi(s, a)^{\top} \mu\left(s^{\prime}\right)\right]\\
&+\mathbb{E}_{(s, a) \sim \mathcal{B}}\left[\int_{\mathcal{S}}\left(\phi(s, a)^{\top} \mu\left(s^{\prime}\right)\right)^2 \mathrm{~d} s^{\prime}\right],
\end{aligned}\label{eq:sample-repr}
\end{equation}
where $C$ is a constant independent of $\simulator\phi,\simulator\mu$. For practical implementations, we can parameterize the $\simulator\phi,\simulator\mu$ both as neural networks (with matching output layer dimensions) and doing gradient descent on $L_{\text{feat}}$.
%problem to learn the representation by optimizing
%\begin{equation}
%	\min_{\simulator{\phi}, \simulator{\mu}} \mathbb{E}_{(s, a, s')\sim \rho\times\simulatorp(\cdot\mid s, a)}\left[\log (\langle\simulator{\phi}(s, a), \simulator{\mu}(s')\rangle)\right]\ .
%\end{equation}
%where $\rho$ is a give state-action pair data distribution with support on $\mathcal{S}\times\mathcal{A}$. Other methods

\begin{figure*}[h]
    \centering
    \includegraphics[width=\linewidth]{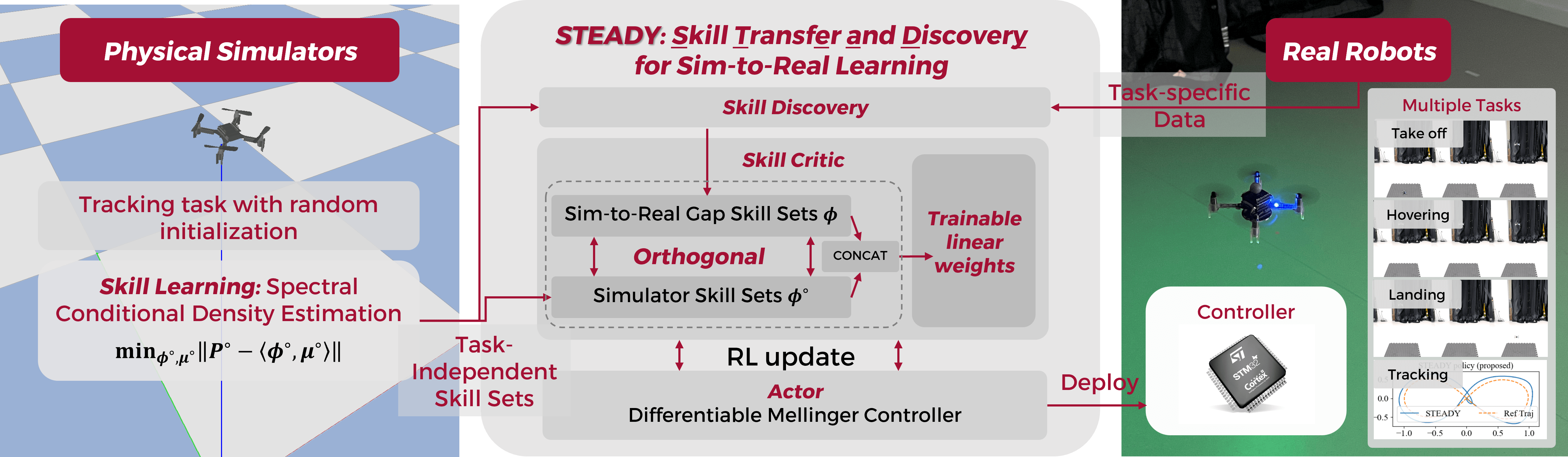}
    \caption{Overview of the \textbf{STEADY} framework for sim-to-real learning. More information can be found on our \href{https://congharvard.github.io/steady-sim-to-real/}{project website}.}
    \label{fig:structure}
\end{figure*}

\section{STEADY: Skill Transfer and Discovery for Sim-to-Real Learning}

In the following two sections, we introduce our methodology of skill transfer and discovery \textbf{(STEADY)} for sim-to-real learning. The whole process includes the following steps: 
\begin{itemize}%[leftmargin=6pt]
    \item[(i)] Learning skill sets and policy in the simulator; 
    % \bo{do we need to learn a policy from simulator? I believe so, since you exploit that in the KL divergnce. }
    \item[(ii)] Discovery of new skills from real-world data;
    \item[(iii)] Policy synthesis from skill sets. 
\end{itemize}  
An overview of the whole framework is shown in Figure \ref{fig:structure}.

\subsection{Learning Skill Sets in the Simulator}
We can leverage the previously mentioned representation learning techniques to learn the simulator skill sets $\simulator{\phi}$ and $\simulator{\mu}$, as well as policy upon the skill sets. In this paper, we follow a similar procedure to the spectral decomposition representation learning (SPEDER) algorithm in \cite{ren_spectral_2023}, which is summarized in Algorithm \ref{alg:simulator}.

\begin{algorithm}
    \caption{Spectral Decomposition Representation \textbf{(SPEDER)} for reinforcement learning \cite{ren_spectral_2023}\label{alg:simulator}} 
    \begin{algorithmic}[1]
        \REQUIRE {simulator MDP $\simulator{\mathcal{M}}$}
        \FOR{episode $n=1,2,\dots,N$}
        \STATE Collect transitions $(s,a,r,s')$ and updates buffer $\mathcal{B}$
        \STATE Learn representation $\simulator{\phi}$ by minimizing \eqref{eq:sample-repr}
        \STATE Update Q function by equation \eqref{eq:pev}
        \STATE Update policy by \eqref{eq:policy_phi}
        \ENDFOR
        \ENSURE simulator policy $\simulator\pi$, representations $\simulator\phi,\simulator\mu$, parameters $w^{\simulator\pi}$
    \end{algorithmic}
\end{algorithm}

\subsection{Skill Discovery from Real-World Data}
After the simulator stage, we transfer learned simulator representations/skill sets, namely $\simulator\phi$, to real robots. However, simply applying the policies $\simulator\pi$ learned from the simulator to the real world, like in zero-shot sim-to-real transfer, might lead to problems due to the sim-to-real gap. Therefore, the major motivation for skill discovery is learning new skills induced by the sim-to-real gap. 

Following a similar procedure of assuming the decomposition structure on $\simulatorp$, we can assume that the sim-to-real gap also admits a spectral decomposition.
\begin{assumption}[Spectral decomposition of sim-to-real gap]
    There exists $\augphi:\mathcal{S}\times\mathcal{A}\to\mathbb R^\orthdim$ and $\augmu:\mathcal{S}\to\mathbb R^\orthdim$ such that
    \begin{equation}
        P(s'|s, a) - \simulatorp(s'|s,a) = \innerprod{\augphi(s,a)}{\augmu(s')} .
    \end{equation}
\end{assumption}

The spectral decomposition allows us to efficiently learn the gap from expensive real-world data, and then a good policy (and potentially more realistic simulators)  given current knowledge of the existing simulator. We then learn the residual dynamics by formulating and solving the following least-square style optimization problem.
\begin{equation}
    \min_{\phi, \mu} \mathbb{E}_{(s, a) \sim \rho_0}\left\|P(\cdot \mid s, a) - \simulatorp(s'\mid s, a) - \innerprod{\augphi(s, a)}{\augmu(s')}\right\|_2^2.\label{eq:s2r_decompose}
\end{equation}

\noindent\textbf{Enforce Skill Discovery by Constraints.} 
Before we proceed to solve the \eqref{eq:s2r_decompose}, % we need additional constraints to make the learning process more sample-efficient. 
% First, we need to make sure that $P$ is a valid probability measure, which means we should make $\int_{\mathcal{S}}P(s'\mid s, a) - \simulatorp(s'\mid s, a)ds'=0$, which adds the constraints
% \begin{equation}
% 	\int_{s'} \augphi^\top(s,a){\augmu}(s') = 0,\quad \forall (s,a) \label{eq: cstr_int}
% \end{equation}
% Moreover, 
we need to make sure that we are \emph{discovering new skills}, which means that the skills learned from the real-world data are different from the previous simulator skills. In mathematical terms, having different skills means that the newly learned representations must be \emph{linearly independent} with all the previous representations. Otherwise, we could simply represent the new skills by a linear combination of the previous skills. In such an undesirable case, the new skills will be redundant when we use the representations to linearly represent the $Q$-function in \eqref{eq:linear_q}. To prevent this and enforce linear independence between the new and simulator skill sets, we add the following orthogonal constraints
% Remember that from the function basis viewpoint in Remark \ref{remark:space}, $\simulator{\phi}$ can be regarded as a set of functional bases of the Q-function $Q^\pi_{\simulatorp}$ induced by arbitrary policy $\pi$, and we only need to fit the linear weights $\theta$ when solving the policy evaluation. 
\begin{equation}
	\innerprod{\simulator{\phi}_i}{{\augphi}_j} = 0,\forall i \in\{1,2,\dots,d\}\text{ and } j\in\{1,2,\dots,s\}\label{eq:orth_cstr}
\end{equation}
where the inner product is defined as
$$
\innerprod{\simulator{\phi}_i}{{\augphi}_j} = \mathbb E_{(s,a)\sim\mathcal{B}}\left[\simulator{\phi}_i(s, a){\augphi}_j(s, a)\right]
$$
The orthogonal constraints in \eqref{eq:orth_cstr} enforce the linear independence between simulator skill sets $\simulator\phi$ and newly learned skill sets $\phi$. Attaching constraints \eqref{eq:orth_cstr}, we have the following optimization problem for skill discovery: 
\begin{equation}
	\begin{aligned}
		\min_{\phi, \mu}& \mathbb{E}_{(s, a) \sim \rho_0}\left\|P(\cdot \mid s, a) - \simulatorp(\cdot\mid s, a)-\augphi(s, a)^{\top} {\augmu}(\cdot)\right\|_2^2 \\
		\text{s.t.} ~& \innerprod{\simulator{\phi}_i}{{\augphi}_j} = 0,\forall i \in\{1,2,\dots,d\}, \ \forall j \in \{1,2,\dots,s\}.
	\end{aligned}
	\label{eq:loss_online}
\end{equation}

\noindent\textbf{Practical Implementations.} 
For the practical implementation of the skill discovery in \eqref{eq:loss_online}, the minimization problem (ignoring the constraints) is similar to the problem in \eqref{eq:least_square_simp}, but apart from estimating $P$, we also need to estimate $\simulatorp$. For $\simulatorp$, we can use the learned representation $\innerprod{\simulator\phi}{\simulator\mu}$ to replace $\simulatorp$. Then we follow a similar idea in \eqref{eq:sample-repr} to minimize 
{
\begin{equation}
    \begin{aligned}
        L_{\text{disc}}(\augphi, \augmu)&:=  \mathbb{E}_{(s, a) \sim \rho_0}\left\|P(\cdot \mid s, a) - \begin{bmatrix}
    \simulator \phi(s, a)\\
    \augphi(s, a)
\end{bmatrix}^{\top} \begin{bmatrix}
    \simulator\mu (\cdot)\\
    {\augmu}(\cdot)
\end{bmatrix}\right\|_2^2 \\
 & = C-2 \mathbb{E}_{(s, a) \sim \mathcal{B}, s^{\prime} \sim P\left(s^{\prime} \mid s, a\right)}\left[\begin{bmatrix}
    \simulator \phi(s, a)\\
    \augphi(s, a)
\end{bmatrix}^{\top} \begin{bmatrix}
    \simulator\mu (\cdot)\\
    {\augmu}(\cdot)
\end{bmatrix}\right]\\
&+\mathbb{E}_{(s, a) \sim \mathcal{B}}\left[\int_{\mathcal{S}}\left(\begin{bmatrix}
    \simulator \phi(s, a)\\
    \augphi(s, a)
\end{bmatrix}^{\top} 
\begin{bmatrix}\label{eq:l2_real_p}
    \simulator\mu (\cdot)\\
    {\augmu}(\cdot)
\end{bmatrix}\right)^2 \mathrm{~d} s^{\prime}\right],
    \end{aligned}
\end{equation}
}
where the simulator skill sets $\simulator\phi,\simulator\mu$ are fixed in the skill discovery stage. We transfer the constraints in \eqref{eq:l2_real_p} to soft constraints by penalty methods for training stability, which leads to the following empirical loss function,
\begin{equation}
    \min_{\phi,\mu}L_{\text{disc}}(\augphi, \augmu) + \lambda \sum_{i,j}\left|\innerprod{\simulator{\phi}_i}{{\augphi}_j}\right|,
\label{eq:practical_skill_disc}
\end{equation}
where $\lambda$ is a hyperparameter penalizing the constraint violations.

\subsection{Policy Synthesis for Real-World Tasks}
\noindent\textbf{Real-world Policy Evaluation.}
After we learn the representations $\augphi,\augmu$, we can leverage the augmented skill sets, $[\simulator\phi, \augphi]$, to synthesize the policies for specific real-world tasks by the MDP planning algorithm such as policy iteration. Here, we also follow a similar policy iteration algorithm to the one used in the simulator stage, while changing the skill sets/representations to be $[\simulator\phi, \augphi]$. Similarly, in the policy evaluation stage, we parameterize the $Q$-function by a linear combination of the enlarged skill sets $[\simulator\phi,\phi]$, $Q(s, a;w) = w_1^\top\simulator\phi(s, a) + w_2^\top\augphi(s, a)$. We can minimize the TD error,
\begin{equation}
    \begin{aligned}
        \min_{w_1,w_2} \mathbb E_{(s,a,r,s')\sim\mathcal{B},a'\sim\pi(\cdot\mid s')}[(& r + \gamma \overline Q(s',a') \\
        &- w_1^\top\simulator\phi(s, a) - w_2^\top\phi(s, a))^2] \label{eq:pev2}
    \end{aligned}
\end{equation}
% Note that here the weights $w_1$ on the simulator representation $\simulator\phi$ might be different with the simulator 
Then we can optimize the policy similar to \eqref{eq:policy_phi} with the new linear parameterization of $Q$-function.

\noindent\textbf{Policy Synthesis.}
When initializing the real-world stage, we initialize the policy by simulator policy $\simulator\pi$ and weights $w_1$ by the simulator learned weights $w^{\simulator\pi}$ so that we do not need to learn the $Q$-function from scratch.
The skill discovery and policy synthesis are conducted simultaneously, and the algorithm is listed in Algorithm \ref{alg:main_alg}. 
To avoid the instability caused by changing from the simulator to the real world, we penalize the $KL$-divergence between the simulator policy and the updated policy to \eqref{eq:policy_phi} when solving the real-world policy improvement,
\begin{equation}
   \max_\pi~ \mathbb{E}_{s\sim \mathcal{B},a\sim\pi(\cdot|s)}[Q(s, a)]-\tau_\pi \bar{D}_{K L}\left(\pi \| \simulator\pi\right)\label{eq:real_pim}
\end{equation}
where $\bar{D}_{K L}\left(\pi \| \simulator\pi\right)=\mathbb{E}_s\left[D_{K L}\left(\pi(\cdot \mid s) \| \simulator\pi(\cdot \mid s)\right)\right]$, $\tau_\pi $ is a hyperparameter penalizing policy update.

\noindent\textbf{Practical Implementations.}
In practical implementations, we add several modifications to the soft actor-critic (SAC) \cite{haarnoja_soft_2018} algorithm. First, we change the $Q$-function parameterization to the linear representation $w_1^\top\simulator\phi(s, a) + w_2^\top\augphi(s, a)$. The representations $\augphi,\augmu, \simulator\phi,\simulator\mu$ are all parameterized by fully-connected neural networks. Second, we use a special parameterization called differentiable Mellinger controllers to parameterize the mean of stochastic actor. The differentiable Mellinger controllers are inspired by the commonly used Mellinger controller for quadrotors \cite{mellinger_minimum_2011}, which has great robustness on real Crazyflies. Details can be found in Section \ref{sec:exp}. The Mellinger controller is a deterministic policy. We parameterized the variance of the stochastic actor using a neural network. The variance network will not be transferred to Crazyflies.
\begin{algorithm}[htb]
	\caption{\textbf{STEADY:} Skill Transfer and Discovery for Sim-to-Real Learning}\label{alg:main_alg}
	\begin{algorithmic}[1]
		\REQUIRE{Simulator MDP $\simulator{\mathcal{M}}$, real world MDP $\mathcal{M}$}
		\STATE \texttt{\# Simulator stage.}
		\STATE $\simulator{\phi}, \simulator{\mu}, \simulator\pi,w^{\simulator\pi}$ = \textbf{SPEDER}($\simulator{\mathcal M}$) (See Algorithm \ref{alg:simulator})
		\STATE \texttt{\# Real-world stage.}
        \STATE Initialization: $\pi^0 = \simulator\pi,w_1^0=w^{\simulator\pi}$
		\FOR{episode $k=1,2,\dots, N$    }
		\STATE Collect trajectory of $\left(s, a, s^{\prime}\right)$ in real world, following current policy $\pi^{k-1}$.
		
		\STATE Skill discovery to learn $\augphi, \augmu$ by optimizing \eqref{eq:practical_skill_disc}, given simulator skills $\simulator\phi,\simulator\mu$. \label{line:skill_discovery}
		% \STATE Exploration strategy TBD
		\STATE Policy evaluation by solving Eq. \eqref{eq:pev2}.
		\STATE Policy improvement by doing policy gradient with respect to \eqref{eq:real_pim}.
		\ENDFOR
            \ENSURE {Final policy $\pi^N$}
            % \vspace{-2mm}
	\end{algorithmic}
 % \vspace{-2mm}
\end{algorithm}
\vspace{-10pt}
%
% \begin{remark}
%     Even though we can linear represent the Q functions for any policy $\pi$ given the linear MDP assumption, we cannot linearly represent the policies or controllers $\pi$. Therefore, we still need to parameterize the $\pi$ as a neural network and use all the deep RL tricks to do the policy improvement empirically.
% \end{remark}

% \begin{remark}[Functional space bases viewpoint.]
%     Another viewpoint of this skills discovery can be the functional bases. For any policy $\pi$, The Q function $Q^pi_{\simulatorp}(s, a)$ lies in the function space $\operatorname{span}\{\simulator\phi_1,\simulator\phi_2,\dots,\simulator\phi_d\}$. Then the skill discovery problem solves bases orthogonal to all bases in $\simulatorp$, which means we have a larger set of the function bases, $\{\simulator{\phi},\phi\}$. Then we optimize the policy by planning on the larger function spaces spanned by $\{\simulator{\phi},\phi\}$.
% \end{remark}

% \begin{equation}
% 	\begin{aligned}
% 		& \min_{w,\tilde w}\left\|r+PV(s') -{w^\pi}^\top\simulator{\phi}(s,a) - \tilde {w^\pi}^\top\augphi(s,a)\right\|\\
% 	\end{aligned}\label{eq:value}
% \end{equation}

% \textbf{Practical Implementation.} Any off-policy RL algorithms could be used to solve \eqref{eq:value} with modifying the critic loss part. In the experimenal section followed in this paper, we use the soft actor-critic (SAC) algorithm, similar to several previous studies \todo{Cite some of Bo's paper}.

\section{Experiments}
\begin{figure*}[htb]
    \centering
    \subfigure[Simulator policy.\label{subfig:takeoff_simulator}]{\includegraphics[width=\linewidth]{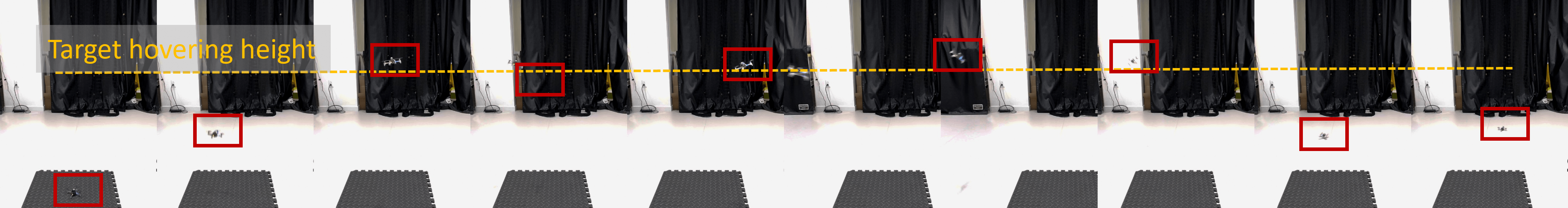}}
    \subfigure[Policy improved by \textbf{STEADY} after training with 20 taking-off and landing trajectories.\label{subfig:takeoff_real}]{\includegraphics[width=\linewidth]{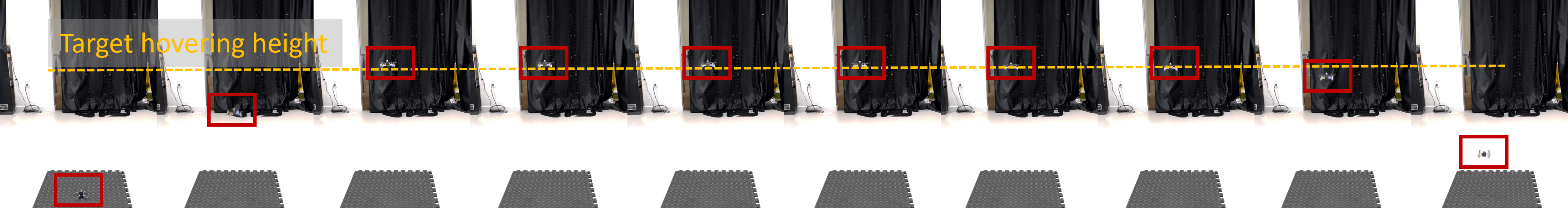}}
    \caption{Snapshots of taking-off, hovering 7 seconds then landing with the simulator policy and policy improved by . Yellow dash lines indicate the target hovering height (1m). The Crazyflies are highlighted with red boxes. The snapshots are taken every 0.8 seconds. Figure \ref{subfig:takeoff_simulator} shows the simulator policy and Figure \ref{subfig:takeoff_real} shows the policy learned by the proposed \textbf{STEADY} algorithm.}
    \label{fig:take_off_land_snapshots}
    \vspace{-4mm}
\end{figure*}

\subsection{Experimental Setup}\label{sec:exp}
\begin{figure}[h]
    \centering
    \subfigure[Experiment environment.\label{fig:env}]{\includegraphics[width=0.45\linewidth]{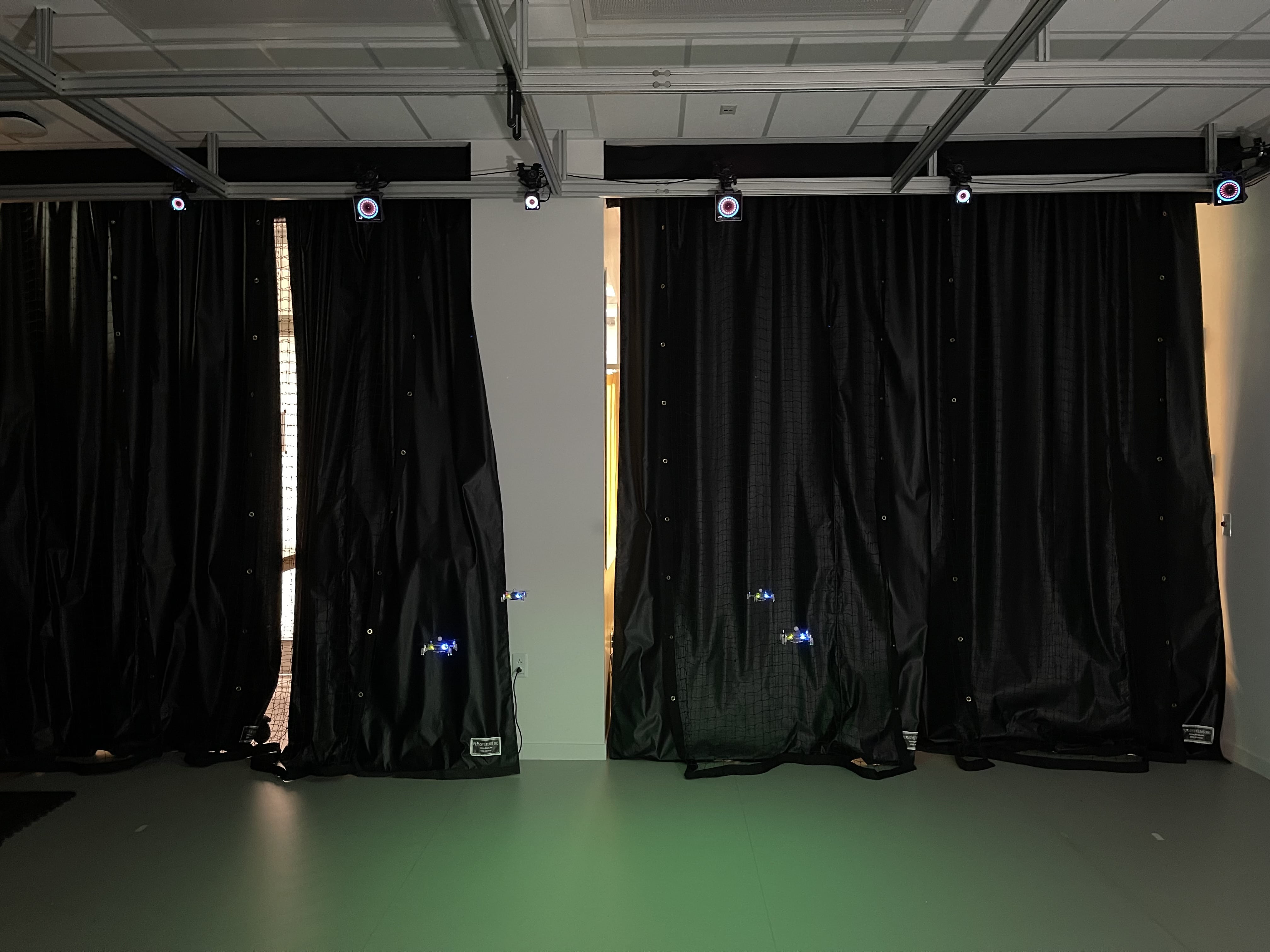}}
    \subfigure[Data recording setup.\label{fig:sdcard}]{\includegraphics[width=0.45\linewidth]{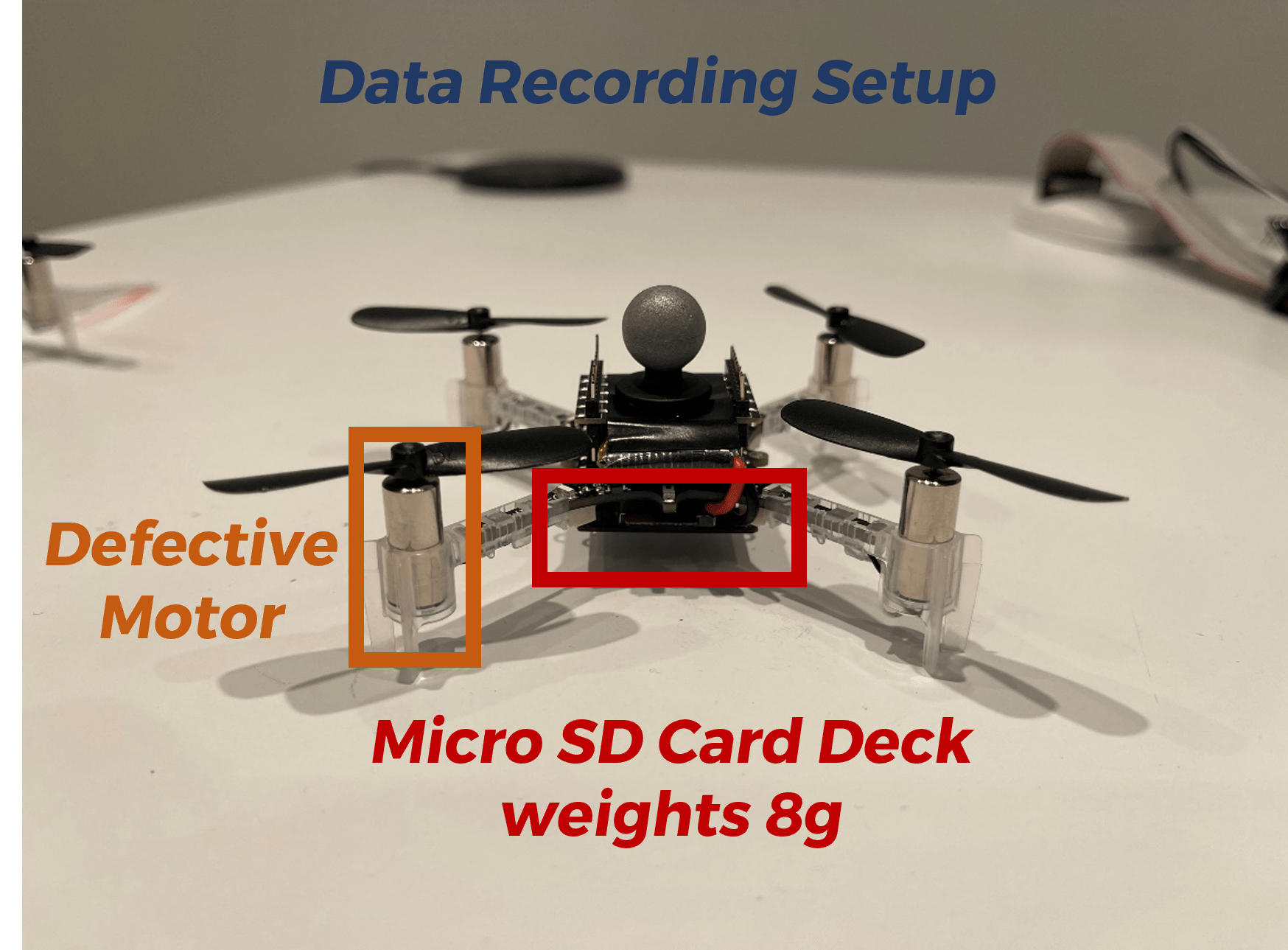}}
    \subfigure[LED highlight setup.\label{fig:ringdeck}]{\includegraphics[width=0.45\linewidth]{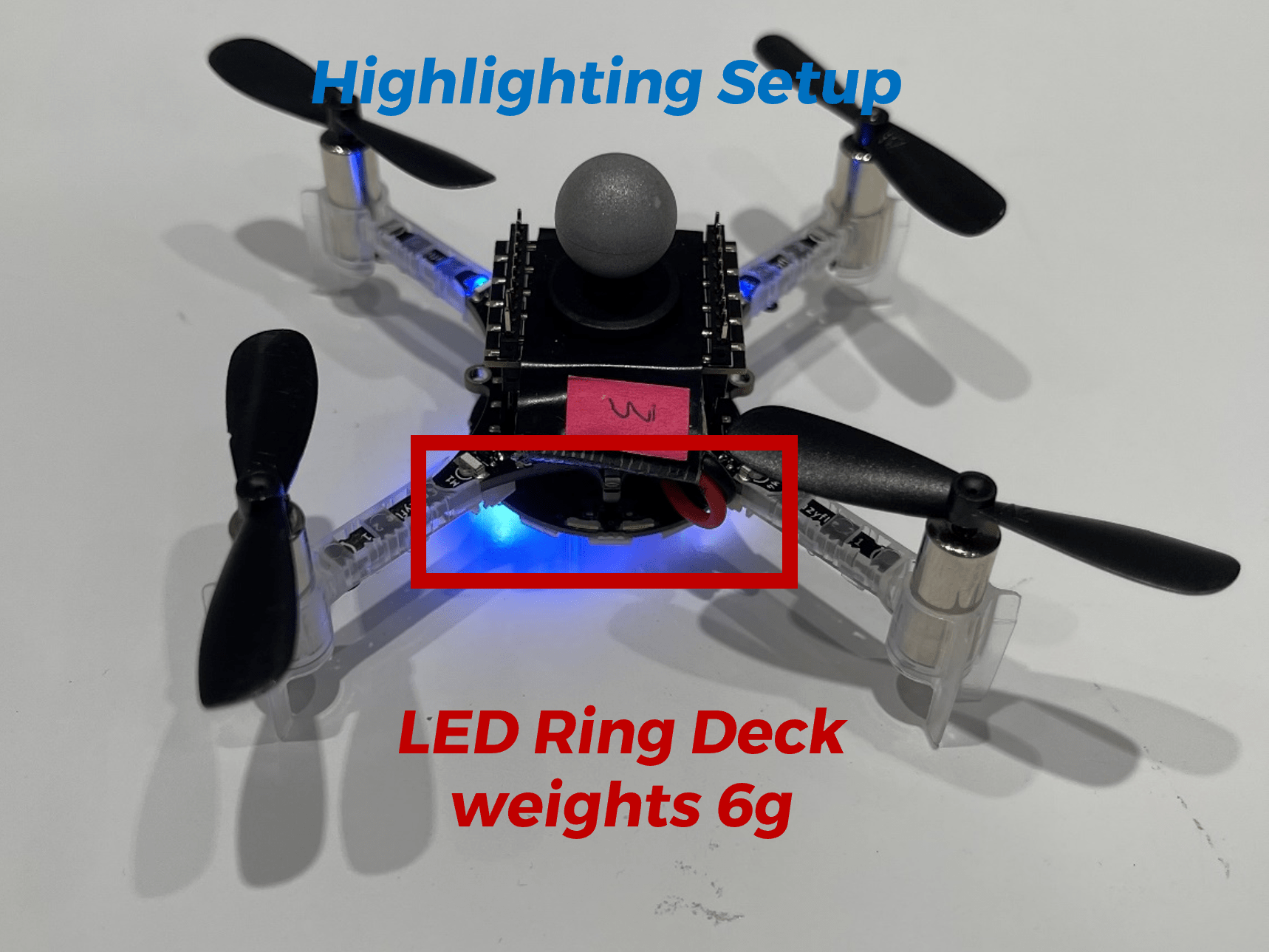}}
    \subfigure[Motor PWM inputs during hovering.\label{fig:motor}]{\includegraphics[width=0.48\linewidth]{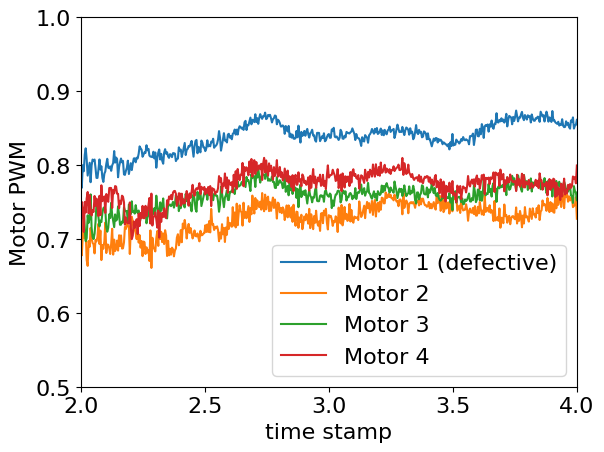}}
    
    \caption{The experiment environment with Optitrack motion capture system.}
    \label{fig:exp}
    \vspace{-2mm}
\end{figure}
We conduct experiments on the quadrotors. Quadrotor is a representative agile and safety-critical dynamical system, which is sensitive to controller malfunctions. Controller failure will immediately cause noticeable oscillations or even falling from the air. Moreover, there are multiple sim-to-real gaps in the quadrotor dynamics that are difficult to model, like motor response \cite{molchanov_sim--multi-real_2019,eschmann_learning_2023} and aerodynamics including ground effect and downwash \cite{shi_neural_2019,shi_neural-swarm_2020}. Therefore, quadrotors are perfect for demonstrating the sim-to-real transfer, which has been widely used in the sim-to-real transfer experiments \cite{kaufmann_champion-level_2023,molchanov_sim--multi-real_2019,gronauer_comparing_2023,shi_neural_2019,o2022neural}.
% We first train the controllers on the simulator, then we deploy the neural-network-based controllers onboard to the quadcopters, finally we collect the data and improve the controllers.

\textbf{Policy Parameterization.} Our transfer learning approaches focus on representing the value function, which supports arbitrary policy parameterization. Although neural-network-based controllers have been implemented on the Crazyflies \cite{molchanov_sim--multi-real_2019,eschmann_learning_2023,gronauer_comparing_2023}, we found that they are usually unstable and not robust to external disturbance such as observation noise, response delays, motor dynamics, etc. To improve controller stability and make the experiments more reproducible, we use a differentiable version of Mellinger controller \cite{mellinger_minimum_2011} as our policy parameterization. The Mellinger controller can be regarded as a modified hierarchical PID controller, which is one of the most commonly used feedback controllers for quadrotors. The policy gradients are applied on the parameters of the Mellinger controllers, which can be found \href{https://www.bitcraze.io/documentation/repository/crazyflie-firmware/master/api/params/#ctrlmel}{here}. Major parameters of Mellinger controller include the PID gain of position, rotation and angular rate error. 

\textbf{Simulator and real-word tasks. }We first train the policies in the simulator for general trajectory tracking tasks. We fix a goal position and randomly initialize the drone states (position, rotation, velocity, angular and positional velocity, propeller rpm) to track the reference. Then for the downstream real-world transfer stage, we consider three different tasks, (1) taking off, hovering, and landing; (2) following a trajectory in the shape of ``8''. We will show that the skill sets learned from the general trajectory tracking tasks are transferable to these specific tasks, and the skill discovery will improve the real-world controller performance.

\textbf{Baselines and ablation studies.} To demonstrate the effectiveness of the proposed controller, we show the comparison between the controller learned by the Algorithm \ref{alg:main_alg} and baselines and ablations. The baseline is the built-in controller in the Crazyflie firmware (labeled as ``Built-in''). Ablation studies including simulator policy (labeled as ``Simulator'') and skill transfer policy (labeled as ``Skill Transfer''). Simulator policy means that we directly use the controller learned in the simulator by Algorithm \ref{alg:simulator}. For the skill transfer policy, we only synthesize policies from the simulator skill sets. The skill discovery in line \ref{line:skill_discovery} of Algorithm \ref{alg:main_alg} is skipped, and the $Q$-function parameterization is the same as the simulator stage. 

\subsection{Simulators Setup}
The dynamical systems of a quadcopter can be regarded as a 3D motion of a rigid body, with four propellers placed at each corner generating force and torque. We use the \texttt{gym-pybullet-drones} simulator \cite{panerati_learning_2021} based on the bullet physical simulator. The simulator frequency and control frequency is both 240 Hz, and the maximum episode length is 480 steps, which is 2s in simulation time.

\textbf{Observations, actions, and reward designs.}
The observations of the system include relative position ${\bf p-p_{goal}}$, rotations/altitude (quaternion ${\bf q}$ and roll, pitch, yaw angles $\{\rm r, p, y\}$),   velocity ${\bf v}$, angular velocity/altitude rate ${\bf \omega}$, and last-step inputs $u_{\text{last}}$, integral and differences of position, altitude, and altitude rate error, rotor PWM inputs. Note that we add the integral and difference of position, altitude, and altitude rate error because the Mellinger controller uses this information.
% To properly handle the symmetry of propeller inputs, we redesign the actions to be desired vertical thrust $F_z$ and rotational forces $F_r,F_p,F_y$. Then these desired forces are converted to desired propeller forces by the power distribution rules
% {\small
% $$
%   \begin{cases}
%    F_1 = F_z - F_r/2 + F_p/2+F_y\\
%    F_2 = F_z - F_r/2 - F_p/2-F_y\\
%    F_3 = F_z + F_r/2 - F_p/2+F_y\\
%    F_4 = F_z + F_r/2 + F_p/2-F_y\\
%   \end{cases}
%   $$
% }
% where $F_i$ is the desired force of $i^{\rm th}$ propeller. 
The reward $r$ is designed as
\begin{equation}
    \begin{aligned}
    r(s, a) = &2 - 2.5 \|{\bf p-p_{{\rm goal}}}\| - 1.5||[{\rm r,p}]|| \\
    &- 0.05 \|{\bf v}\| - 0.05\|\omega\|-0.1\|u\|  \\% -c_{{\rm actdiff}}\|u-u_{\text{last}}\|
\end{aligned}\label{eq:rew}
\end{equation}
The initial distribution and other detailed simulator settings can be found in Appendix \ref{appendix:simu} of our online report \cite{online_report}.

\noindent\textbf{Domain randomization and curriculum learning.} Some existing sim-to-real transfer on quadrotors have used domain randomization\cite{molchanov_sim--multi-real_2019} on the physical parameters and used curriculum learning to speed up the training \cite{eschmann_learning_2023}. Curriculum learning means adaptively changing the reward weights in \eqref{eq:rew} to improve performance. These aspects are not the major focus of our paper, and for crazyflie the physical parameters has been identified by existing papers. Therefore, we did not add these methodologies to our simulators. 

\subsection{Real-World Robots Setup}
We show the effectiveness of our sim-to-real learning on the Crazyflie 2.1 Quadrotor with an STM32F405 microcontroller clocked at 168 MHz. 
We transfer the learned controller parameters to Crazyflie and handle the communication using Crazyswarm \cite{preiss_crazyswarm_2017}. We use the Optitrack motion capture system with single-marker configuration to provide location information for the Crazyflie, shown in Figure \ref{fig:env}. The Crazyflie carries an additional micro SD card deck to record data or a LED ring deck to highlight itself shown in Figure \ref{fig:sdcard} and \ref{fig:ringdeck}, respectively.

\textbf{Causes of sim-to-real gaps.} The sim-to-real gaps include several factors. (1) Extra weight from the decks carried. The additional decks weigh 6-8g each, which is a noticeable weight change compared to the original weight of Crazyflie, 27g. (2) For the taking-off, hovering, and landing tasks, we found out that our Crazyflie used for these specific tasks has a defective motor that requires higher PWM input to balance shown in Figure \ref{fig:motor}. (3) Other sim-to-real gaps that are not captured by models like motion capture noise and network fluctuations.

% \vspace{-2mm}
\subsection{Experimental Results}
\subsubsection{Simulation Stage}
\begin{figure}[htbp]
    \centering
    \vspace{-10pt}
    \includegraphics[width=0.6\linewidth]{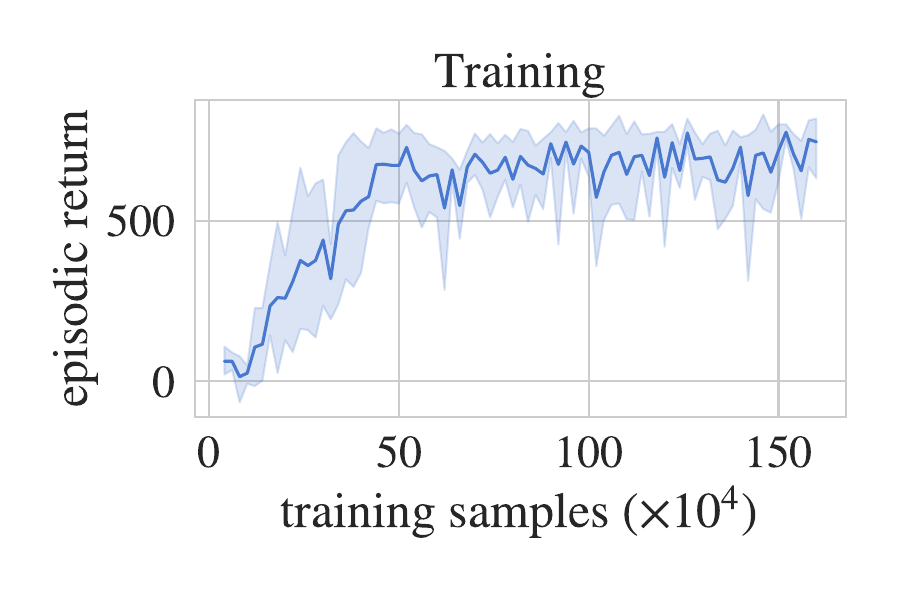}
    \vspace{-4mm}
    \caption{Episodic return to training samples during the training process with 5 random seeds. The shaded region implies 95\% confidential interval over 10 evaluation episodes for every 2000 samples of the five random seeds.}
    \label{fig:training}
\end{figure}

The features $\simulator\phi, \phi,\simulator\mu, \mu$ are parameterized by fully connected neural networks with two hidden layers with 256 neurons. The dimensions of all the representations are 256. For the simulation stage, we train the algorithm with 1.6$\times$10\textsuperscript{6} transitions and the expected return during the training process are shown in Figure \ref{fig:training}.
\vspace{-10pt}
% \begin{figure}[htb]
%     \centering
%     \includegraphics[width=0.45\linewidth]{fig_exp/sdcard.png}
%     \includegraphics[width=0.45\linewidth]{fig/defective.png}
%     \caption{Crazyflie and its motor PWM inputs for hovering. Motor 1 is defective which requires higher PWM input to keep hovering.}
%     \label{fig:crazyflie}
% \end{figure}

\subsubsection{Real-world Stage}
\begin{figure}[h]
    \centering
    \vspace{-4mm}
    \includegraphics[width=\linewidth]{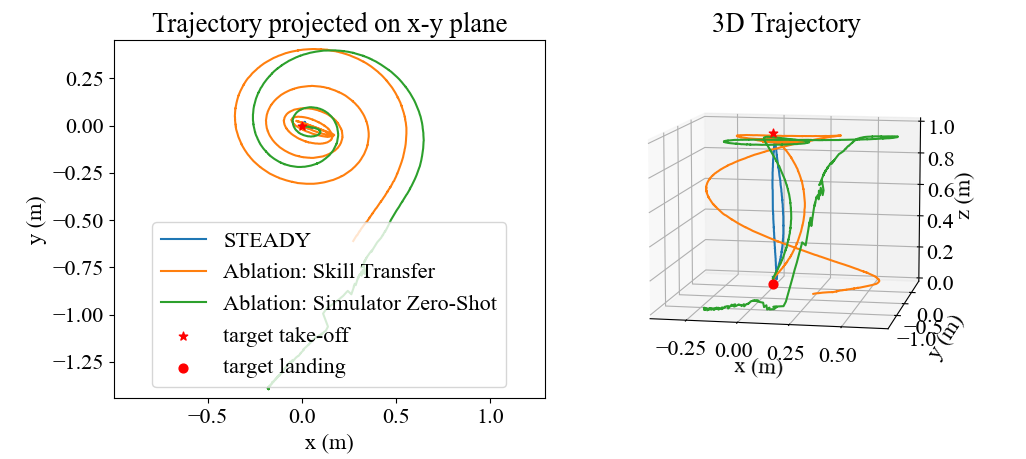}
    \caption{Trajectories of taking-off, hovering and landing trajectories.}
    \label{fig:3dtraj_takeoff}
    \vspace{-2mm}
\end{figure}

\begin{figure}[h]
    \centering
    \includegraphics[width=\linewidth]{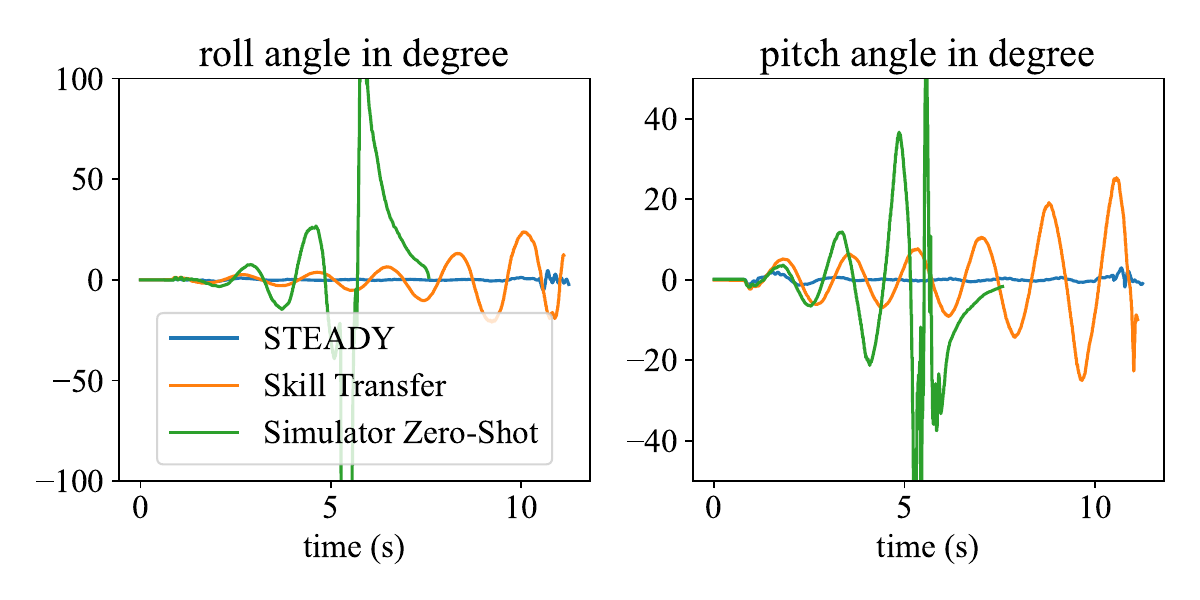}
    \caption{Roll and pitch angles during taking off, hovering and landing experiments.}
    \label{fig:rollpitch_takeoff}
    \vspace{-2mm}
\end{figure}

\textbf{Taking off, hovering and landing.} First we show the experimental results on the real-world task of taking off, hovering for 7 seconds at 1m and landing. We compare the 3D trajectories in Figure \ref{fig:3dtraj_takeoff} and roll and pitch angle in Figure \ref{fig:rollpitch_takeoff} with skill transfer and discovery policy (labeled as \textbf{``STEADY''}), skill transfer policy (labeled as \textbf{``Skill Tranfer''}) and simulator policy (labeled as \textbf{``Simulator Zero-Shot''}, Zero-Shot means that no learning from real-world data is implemented.). Two sequences of snapshots comparing Simulator policy and the policy improved by STEADY are also shown in Figure \ref{fig:take_off_land_snapshots}, and the full video can be found at \href{https://congharvard.github.io/steady-sim-to-real/}{https://congharvard.github.io/steady-sim-to-real/}. Results show that after learning from the real-world data using the STEADY framework, the controller can maintain at the target height very stably without oscillations even with a defective motor and extra weight carried. The stability of the STEADY controller can also be verified from the flat roll and pitch angle curves in Figure \ref{fig:rollpitch_takeoff}. For the controllers from ablation studies, the simulator policy oscillate heavily and cannot maintain the height. The skill transfer policy stay in the air longer than the simulator policy but still fails to maintain stably at the target position.

\begin{figure}[h]
    \centering
    \vspace{-5mm}
    \includegraphics[width=\linewidth]{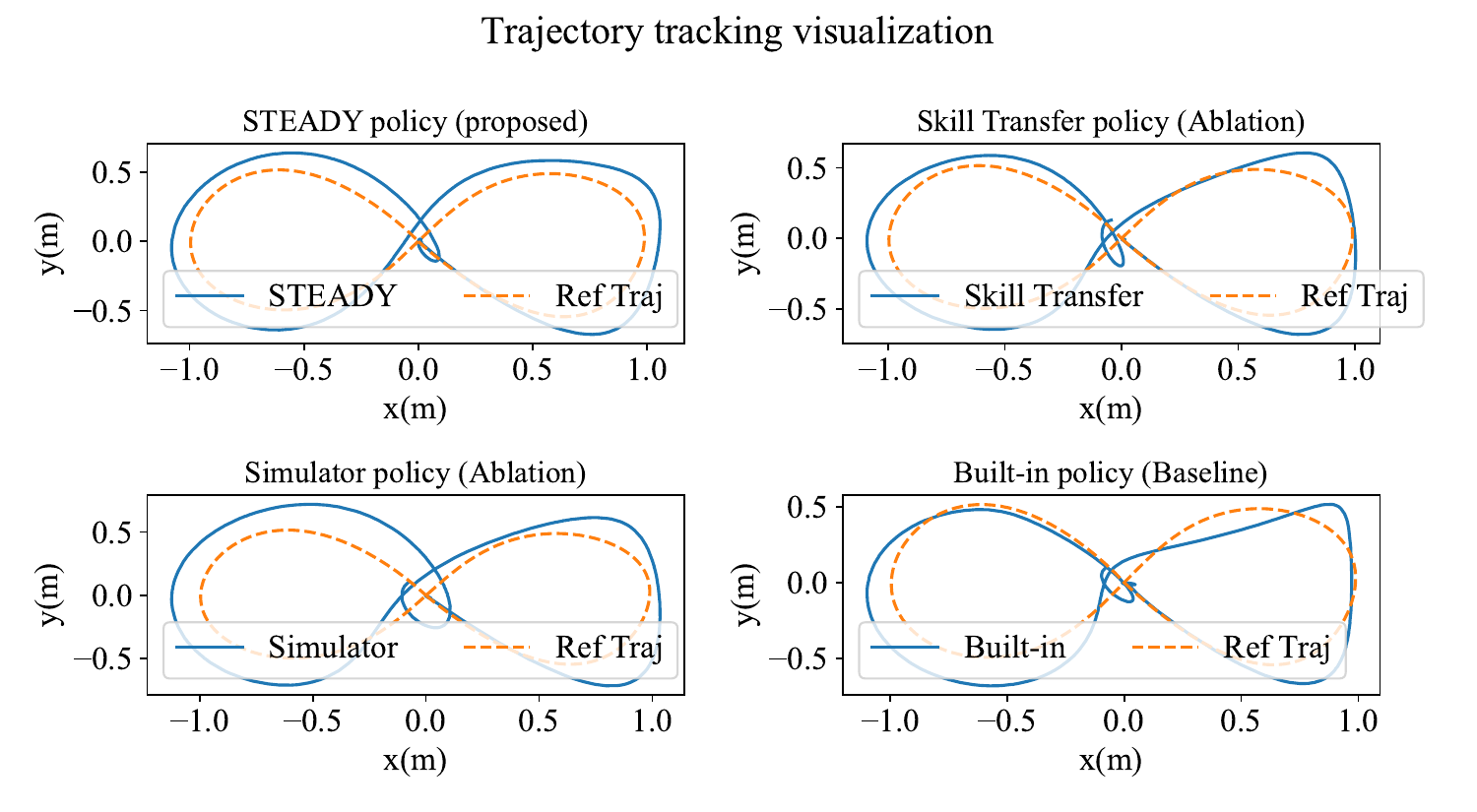}
    \vspace{-6mm}
    \caption{Trajectory tracking visualization.}
    \label{fig:tracking_fig8}
\end{figure}

\textbf{Trajectory tracking.} For the trajectory tracking task, the drone needs to follow a trajectory in the shape of the number ``8'', which is commonly seen in related studies \cite{o2022neural,eschmann_learning_2023}. The trajectories is $(x(t),y(t),z(t))=(\sin(t),\frac{1}{2}\sin(2t), 1.0)$ and the shape is shown as orange dash lines in Figure \ref{fig:tracking_fig8}. Figure \ref{fig:tracking_fig8} shows the trajectory tracking results with different controllers compared to reference trajectories. The performance is compared in Table \ref{tab:tracking_perf}. We calculate two metrics for the tracking error, the average position tracking error and the cumulative rewards. The average position tracking error averages $\|{\bf p - p_{\text{goal}}}\|$ over the trajectory. The cumulative rewards sums up rewards defined in \eqref{eq:rew} over the trajectory (removing the initial constant term of $2$). Table \ref{tab:tracking_perf} shows that the STEADY controller achieves the smallest trajectory tracking error and highest accumulative rewards. The tracking error is comparable to the built-in controllers and improved by 11.9\% and 30.2\% compared to ablation skill transfer controller and ablation simulator controller, respectively. For the cumulative reward, the STEADY controller outperforms all three baseline controllers, improving on the ablation skill transfer controller by 9.1\%, the Built-in PID controller by 10.9\%, and the ablation simulator controller by 22.7\%. 

\begin{table}[h]
\centering
{\small
\vspace{-8mm}
\caption{Tracking performance comparison.} \label{tab:tracking_perf}
\begin{tabular}{@{}lcc@{}}
\toprule
                & Average Tracking          & Cumulative \\
Controller      & Error ($\times 10^{-1}$m) &  Rewards \\\midrule
\textbf{STEADY} & \textbf{0.982}            & \textbf{-1241}            \\
Skill Transfer  & 1.115                     & -1365            \\
Simulator       & 1.406                     & -1584            \\
Built-in PID    & 0.983                     & -1392            \\ \bottomrule
\end{tabular}
}
\vspace{-6mm}
\end{table}

\section{Concluding Remarks}

In this paper, we proposed the STEADY framework, which utilizes skill transfer and discovery for sim-to-real learning. Inspired by the concept of representations as skill sets when considering the spectral decompositions of MDPs, we show that we can learn the skill sets from simulators and then transfer them to the real world. The representation-based skill sets can also help sim-to-real transfer. By enforcing orthogonal constraints between the simulator skill sets and the skill sets induced by the sim-to-real gap, we promote the discovery of useful and distinct new skills. Building on the enlarged skill sets comprising these new skill sets and the existing simulator skill sets, STEADY facilitates more efficient and effective sim-to-real transfer smoothly. 
% As a work aiming to bridge theoretical and empirical community, this paper still has a lot of limitations, which is discussed in the online report. Future works include addressing these limitations.

% can learn skill transfer more efficiently {\color{blue} need to finish the conclusion by adding a few more words/another sentence}  
\bibliographystyle{format/IEEEtran}
\bibliography{cleaned_ref}

\begin{thebibliography}{10}
\providecommand{\url}[1]{#1}
\csname url@samestyle\endcsname
\providecommand{\newblock}{\relax}
\providecommand{\bibinfo}[2]{#2}
\providecommand{\BIBentrySTDinterwordspacing}{\spaceskip=0pt\relax}
\providecommand{\BIBentryALTinterwordstretchfactor}{4}
\providecommand{\BIBentryALTinterwordspacing}{\spaceskip=\fontdimen2\font plus
\BIBentryALTinterwordstretchfactor\fontdimen3\font minus
  \fontdimen4\font\relax}
\providecommand{\BIBforeignlanguage}[2]{{%
\expandafter\ifx\csname l@#1\endcsname\relax
\typeout{** WARNING: IEEEtran.bst: No hyphenation pattern has been}%
\typeout{** loaded for the language `#1'. Using the pattern for}%
\typeout{** the default language instead.}%
\else
\language=\csname l@#1\endcsname
\fi
#2}}
\providecommand{\BIBdecl}{\relax}
\BIBdecl

\bibitem{silver_mastering_2017}
D.~Silver, J.~Schrittwieser, K.~Simonyan, I.~Antonoglou, A.~Huang, A.~Guez,
  T.~Hubert, L.~Baker, M.~Lai, A.~Bolton, Y.~Chen, T.~Lillicrap, F.~Hui,
  L.~Sifre, G.~van~den Driessche, T.~Graepel, and D.~Hassabis,
  ``\BIBforeignlanguage{en}{Mastering the game of {Go} without human
  knowledge},'' \emph{\BIBforeignlanguage{en}{Nature}}, vol. 550, no. 7676, pp.
  354--359, Oct. 2017, number: 7676 Publisher: Nature Publishing Group.

\bibitem{haarnoja_soft_2018}
T.~Haarnoja, A.~Zhou, P.~Abbeel, and S.~Levine, ``\BIBforeignlanguage{en}{Soft
  {Actor}-{Critic}: {Off}-{Policy} {Maximum} {Entropy} {Deep} {Reinforcement}
  {Learning} with a {Stochastic} {Actor}},'' in
  \emph{\BIBforeignlanguage{en}{Proceedings of the 35th {International}
  {Conference} on {Machine} {Learning}}}.\hskip 1em plus 0.5em minus
  0.4em\relax PMLR, Jul. 2018, pp. 1861--1870.

\bibitem{wurman_outracing_2022}
P.~R. Wurman, S.~Barrett, K.~Kawamoto, J.~MacGlashan, K.~Subramanian, T.~J.
  Walsh, R.~Capobianco, A.~Devlic, F.~Eckert, F.~Fuchs, L.~Gilpin,
  P.~Khandelwal, V.~Kompella, H.~Lin, P.~MacAlpine, D.~Oller, T.~Seno,
  C.~Sherstan, M.~D. Thomure, H.~Aghabozorgi, L.~Barrett, R.~Douglas,
  D.~Whitehead, P.~Dürr, P.~Stone, M.~Spranger, and H.~Kitano,
  ``\BIBforeignlanguage{en}{Outracing champion {Gran} {Turismo} drivers with
  deep reinforcement learning},'' \emph{\BIBforeignlanguage{en}{Nature}}, vol.
  602, no. 7896, pp. 223--228, Feb. 2022, number: 7896 Publisher: Nature
  Publishing Group.

\bibitem{andrychowicz_learning_2020}
O.~M. Andrychowicz, B.~Baker, M.~Chociej, R.~Józefowicz, B.~McGrew,
  J.~Pachocki, A.~Petron, M.~Plappert, G.~Powell, A.~Ray, J.~Schneider,
  S.~Sidor, J.~Tobin, P.~Welinder, L.~Weng, and W.~Zaremba,
  ``\BIBforeignlanguage{en}{Learning dexterous in-hand manipulation},''
  \emph{\BIBforeignlanguage{en}{The International Journal of Robotics
  Research}}, vol.~39, no.~1, pp. 3--20, Jan. 2020, publisher: SAGE
  Publications Ltd STM.

\bibitem{kaufmann_champion-level_2023}
E.~Kaufmann, L.~Bauersfeld, A.~Loquercio, M.~Müller, V.~Koltun, and
  D.~Scaramuzza, ``\BIBforeignlanguage{en}{Champion-level drone racing using
  deep reinforcement learning},'' \emph{\BIBforeignlanguage{en}{Nature}}, vol.
  620, no. 7976, pp. 982--987, Aug. 2023, number: 7976 Publisher: Nature
  Publishing Group.

\bibitem{molchanov_sim--multi-real_2019}
A.~Molchanov, T.~Chen, W.~H{\"o}nig, J.~A. Preiss, N.~Ayanian, and G.~S.
  Sukhatme, ``Sim-to-(multi)-real: Transfer of low-level robust control
  policies to multiple quadrotors,'' in \emph{2019 IEEE/RSJ International
  Conference on Intelligent Robots and Systems (IROS)}.\hskip 1em plus 0.5em
  minus 0.4em\relax IEEE, 2019, pp. 59--66.

\bibitem{gronauer_comparing_2023}
S.~Gronauer, D.~Stümke, and K.~Diepold, ``\BIBforeignlanguage{en}{Comparing
  {Quadrotor} {Control} {Policies} for {Zero}-{Shot} {Reinforcement} {Learning}
  under {Uncertainty} and {Partial} {Observability}},'' in
  \emph{\BIBforeignlanguage{en}{2023 {IEEE}/{RSJ} {International} {Conference}
  on {Intelligent} {Robots} and {Systems} ({IROS})}}.\hskip 1em plus 0.5em
  minus 0.4em\relax Detroit, MI, USA: IEEE, Oct. 2023, pp. 7508--7514.

\bibitem{jin_provably_2019}
C.~Jin, Z.~Yang, Z.~Wang, and M.~I. Jordan, ``Provably efficient reinforcement
  learning with linear function approximation,'' in \emph{Conference on
  learning theory}.\hskip 1em plus 0.5em minus 0.4em\relax PMLR, 2020, pp.
  2137--2143.

\bibitem{yang_reinforcement_2020}
L.~Yang and M.~Wang, ``Reinforcement learning in feature space: Matrix bandit,
  kernels, and regret bound,'' in \emph{International Conference on Machine
  Learning}.\hskip 1em plus 0.5em minus 0.4em\relax PMLR, 2020, pp.
  10\,746--10\,756.

\bibitem{agarwal_flambe_2020}
A.~Agarwal, S.~Kakade, A.~Krishnamurthy, and W.~Sun, ``Flambe: Structural
  complexity and representation learning of low rank mdps,'' \emph{Advances in
  neural information processing systems}, vol.~33, pp. 20\,095--20\,107, 2020.

\bibitem{uehara_representation_2022}
M.~Uehara, X.~Zhang, and W.~Sun, ``Representation learning for online and
  offline rl in low-rank mdps,'' \emph{arXiv preprint arXiv:2110.04652}, 2021.

\bibitem{ren_spectral_2023}
T.~Ren, T.~Zhang, L.~Lee, J.~E. Gonzalez, D.~Schuurmans, and B.~Dai, ``Spectral
  decomposition representation for reinforcement learning,'' in \emph{The
  Eleventh International Conference on Learning Representations}, 2023.

\bibitem{zhang_making_2022}
T.~Zhang, T.~Ren, M.~Yang, J.~Gonzalez, D.~Schuurmans, and B.~Dai, ``Making
  linear mdps practical via contrastive representation learning,'' in
  \emph{International Conference on Machine Learning}.\hskip 1em plus 0.5em
  minus 0.4em\relax PMLR, 2022, pp. 26\,447--26\,466.

\bibitem{ren_latent_2023}
T.~Ren, C.~Xiao, T.~Zhang, N.~Li, Z.~Wang, sujay sanghavi, D.~Schuurmans, and
  B.~Dai, ``Latent variable representation for reinforcement learning,'' in
  \emph{The Eleventh International Conference on Learning Representations},
  2023.

\bibitem{saveriano_data-efficient_2017}
M.~Saveriano, Y.~Yin, P.~Falco, and D.~Lee, ``Data-efficient control policy
  search using residual dynamics learning,'' in \emph{2017 {IEEE}/{RSJ}
  {International} {Conference} on {Intelligent} {Robots} and {Systems}
  ({IROS})}, Sep. 2017, pp. 4709--4715.

\bibitem{johannink_residual_2019}
T.~Johannink, S.~Bahl, A.~Nair, J.~Luo, A.~Kumar, M.~Loskyll, J.~A. Ojea,
  E.~Solowjow, and S.~Levine, ``\BIBforeignlanguage{en}{Residual
  {Reinforcement} {Learning} for {Robot} {Control}},'' in
  \emph{\BIBforeignlanguage{en}{2019 {International} {Conference} on {Robotics}
  and {Automation} ({ICRA})}}.\hskip 1em plus 0.5em minus 0.4em\relax Montreal,
  QC, Canada: IEEE, May 2019, pp. 6023--6029.

\bibitem{shi_neural_2019}
G.~Shi, X.~Shi, M.~O’Connell, R.~Yu, K.~Azizzadenesheli, A.~Anandkumar,
  Y.~Yue, and S.-J. Chung, ``Neural {Lander}: {Stable} {Drone} {Landing}
  {Control} {Using} {Learned} {Dynamics},'' in \emph{2019 {International}
  {Conference} on {Robotics} and {Automation} ({ICRA})}, May 2019, pp.
  9784--9790.

\bibitem{ren_stochastic_2023}
T.~Ren, Z.~Ren, N.~Li, and B.~Dai, ``Stochastic nonlinear control via
  finite-dimensional spectral dynamic embedding,'' in \emph{2023 62nd IEEE
  Conference on Decision and Control (CDC)}.\hskip 1em plus 0.5em minus
  0.4em\relax IEEE, 2023, pp. 795--800.

\bibitem{tobin_domain_2017}
J.~Tobin, R.~Fong, A.~Ray, J.~Schneider, W.~Zaremba, and P.~Abbeel, ``Domain
  randomization for transferring deep neural networks from simulation to the
  real world,'' in \emph{2017 {IEEE}/{RSJ} {International} {Conference} on
  {Intelligent} {Robots} and {Systems} ({IROS})}, Sep. 2017, pp. 23--30.

\bibitem{chebotar_closing_2019}
Y.~Chebotar, A.~Handa, V.~Makoviychuk, M.~Macklin, J.~Issac, N.~Ratliff, and
  D.~Fox, ``Closing the {Sim}-to-{Real} {Loop}: {Adapting} {Simulation}
  {Randomization} with {Real} {World} {Experience},'' in \emph{2019
  {International} {Conference} on {Robotics} and {Automation} ({ICRA})}, May
  2019, pp. 8973--8979.

\bibitem{peng_sim--real_2018}
X.~B. Peng, M.~Andrychowicz, W.~Zaremba, and P.~Abbeel, ``Sim-to-{Real}
  {Transfer} of {Robotic} {Control} with {Dynamics} {Randomization},'' in
  \emph{2018 {IEEE} {International} {Conference} on {Robotics} and {Automation}
  ({ICRA})}, May 2018, pp. 3803--3810.

\bibitem{levine_learning_2014}
S.~Levine and P.~Abbeel, ``Learning {Neural} {Network} {Policies} with {Guided}
  {Policy} {Search} under {Unknown} {Dynamics},'' in \emph{Advances in {Neural}
  {Information} {Processing} {Systems}}, vol.~27.\hskip 1em plus 0.5em minus
  0.4em\relax Curran Associates, Inc., 2014.

\bibitem{shi_neural-swarm_2020}
G.~Shi, W.~Hönig, Y.~Yue, and S.-J. Chung, ``Neural-{Swarm}: {Decentralized}
  {Close}-{Proximity} {Multirotor} {Control} {Using} {Learned}
  {Interactions},'' in \emph{2020 {IEEE} {International} {Conference} on
  {Robotics} and {Automation} ({ICRA})}, May 2020, pp. 3241--3247, iSSN:
  2577-087X.

\bibitem{fisac_general_2019}
J.~F. Fisac, A.~K. Akametalu, M.~N. Zeilinger, S.~Kaynama, J.~Gillula, and
  C.~J. Tomlin, ``A {General} {Safety} {Framework} for {Learning}-{Based}
  {Control} in {Uncertain} {Robotic} {Systems},'' \emph{IEEE Transactions on
  Automatic Control}, vol.~64, no.~7, pp. 2737--2752, Jul. 2019, conference
  Name: IEEE Transactions on Automatic Control.

\bibitem{du_auto-tuned_2021}
Y.~Du, O.~Watkins, T.~Darrell, P.~Abbeel, and D.~Pathak,
  ``\BIBforeignlanguage{en}{Auto-{Tuned} {Sim}-to-{Real} {Transfer}},'' in
  \emph{\BIBforeignlanguage{en}{2021 {IEEE} {International} {Conference} on
  {Robotics} and {Automation} ({ICRA})}}.\hskip 1em plus 0.5em minus
  0.4em\relax Xi'an, China: IEEE, May 2021, pp. 1290--1296.

\bibitem{tanwani_dirl_2021}
A.~Tanwani, ``\BIBforeignlanguage{en}{{DIRL}: {Domain}-{Invariant}
  {Representation} {Learning} for {Sim}-to-{Real} {Transfer}},'' in
  \emph{\BIBforeignlanguage{en}{Proceedings of the 2020 {Conference} on {Robot}
  {Learning}}}.\hskip 1em plus 0.5em minus 0.4em\relax PMLR, Oct. 2021, pp.
  1558--1571.

\bibitem{devin_learning_2017}
C.~Devin, A.~Gupta, T.~Darrell, P.~Abbeel, and S.~Levine, ``Learning modular
  neural network policies for multi-task and multi-robot transfer,'' in
  \emph{2017 {IEEE} {International} {Conference} on {Robotics} and {Automation}
  ({ICRA})}, May 2017, pp. 2169--2176.

\bibitem{helwa_multi-robot_2017}
M.~K. Helwa and A.~P. Schoellig, ``Multi-robot transfer learning: {A} dynamical
  system perspective,'' in \emph{2017 {IEEE}/{RSJ} {International} {Conference}
  on {Intelligent} {Robots} and {Systems} ({IROS})}, Sep. 2017, pp. 4702--4708.

\bibitem{agarwal_provable_2023}
A.~Agarwal, Y.~Song, W.~Sun, K.~Wang, M.~Wang, and X.~Zhang,
  ``\BIBforeignlanguage{en}{Provable {Benefits} of {Representational}
  {Transfer} in {Reinforcement} {Learning}},'' in
  \emph{\BIBforeignlanguage{en}{Proceedings of {Thirty} {Sixth} {Conference} on
  {Learning} {Theory}}}.\hskip 1em plus 0.5em minus 0.4em\relax PMLR, Jul.
  2023, pp. 2114--2187.

\bibitem{mnih_playing_2013}
V.~Mnih, K.~Kavukcuoglu, D.~Silver, A.~A. Rusu, J.~Veness, M.~G. Bellemare,
  A.~Graves, M.~Riedmiller, A.~K. Fidjeland, G.~Ostrovski \emph{et~al.},
  ``Human-level control through deep reinforcement learning,'' \emph{nature},
  vol. 518, no. 7540, pp. 529--533, 2015.

\bibitem{mellinger_minimum_2011}
D.~Mellinger and V.~Kumar, ``Minimum snap trajectory generation and control for
  quadrotors,'' in \emph{2011 {IEEE} {International} {Conference} on {Robotics}
  and {Automation}}.\hskip 1em plus 0.5em minus 0.4em\relax Shanghai, China:
  IEEE, May 2011, pp. 2520--2525.

\bibitem{eschmann_learning_2023}
J.~Eschmann, D.~Albani, and G.~Loianno, ``Learning to fly in seconds,'' 2023.

\bibitem{o2022neural}
M.~O’Connell, G.~Shi, X.~Shi, K.~Azizzadenesheli, A.~Anandkumar, Y.~Yue, and
  S.-J. Chung, ``Neural-fly enables rapid learning for agile flight in strong
  winds,'' \emph{Science Robotics}, vol.~7, no.~66, p. eabm6597, 2022.

\bibitem{panerati_learning_2021}
J.~Panerati, H.~Zheng, S.~Zhou, J.~Xu, A.~Prorok, and A.~P. Schoellig,
  ``Learning to fly—a gym environment with pybullet physics for reinforcement
  learning of multi-agent quadcopter control,'' in \emph{2021 IEEE/RSJ
  International Conference on Intelligent Robots and Systems (IROS)}.\hskip 1em
  plus 0.5em minus 0.4em\relax IEEE, 2021, pp. 7512--7519.

\bibitem{online_report}
\BIBentryALTinterwordspacing
H.~Ma, Z.~Ren, B.~Dai, and N.~Li, ``Skill transfer and discovery for
  sim-to-real learning: A representation-based viewpoint.'' [Online].
  Available:
  \url{https://scholar.harvard.edu/haitongma/files/transfer_learning_online_report.pdf}
\BIBentrySTDinterwordspacing

\bibitem{preiss_crazyswarm_2017}
J.~A. Preiss, W.~Honig, G.~S. Sukhatme, and N.~Ayanian, ``Crazyswarm: A large
  nano-quadcopter swarm,'' in \emph{2017 IEEE International Conference on
  Robotics and Automation (ICRA)}.\hskip 1em plus 0.5em minus 0.4em\relax IEEE,
  2017, pp. 3299--3304.

\end{thebibliography}
\clearpage
\appendices
\section{Detailed Algorithm Implementation}
\label{appendix:algo}

\subsection{Differentiable Mellinger controllers.} 
\textbf{Notations.} In the rigid body dynamics, we need to be careful with the reference frame. We use $W$ to denote the world frame, $B$ to denote the body frame of the crazyflie. We use ${ }^W R_B$ to denote the rotation matrix of frame $B$ with respect to frame $W$. the $x, y, z$ axes of $B$ frame are denoted by $\mathbf{x}_B,\mathbf{y}_B,\mathbf{z}_B$. We also use $Z - X - Y$ Euler angles to define the roll, pitch, and yaw angles ($\phi$, $\theta$, and $\psi$) as a local coordinate system.
The angular velocity of the robot is denoted by $\omega_{{B} W}$, denoting the angular velocity of frame ${B}$ in the frame ${W}$, with components $p, q$, and $r$ in the body frame:
$$
\omega_{\mathcal{B W}}=p \mathbf{x}_B+q \mathbf{y}_B+r \mathbf{z}_B .
$$

\textbf{Mellinger Controller.} The Mellinger controller was proposed in \cite{mellinger_minimum_2011}, which is a modified version of the PID controller. The design is similar to a hierarchical PID controller. We will briefly introduce the Mellinger controller here and then discuss our modification to the Mellinger controller to make it as our policy parameterizations. The Mellinger controller starts from the position and velocity error, 
$$
\mathbf{e}_p=\mathbf{p}-\mathbf{r}_T, \mathbf{e}_v=\mathbf{v}-\dot{\mathbf{r}}_T
$$
where $\mathbf{r}_T$ is the 3D vector of the positions of the reference trajectory. Then we compute the desired force vector by 
$$
\mathbf{F}_{d e s}=-K_p \mathbf{e}_p-K_v \mathbf{e}_v+m g \mathbf{z}_W+m \ddot{\mathbf{r}}_T
$$
where $K_p$ and $K_v$ are positive definite gain matrices. Next we project the desired force to the $z$-axis of the current body frame, which is the first input, desired thrust,
$$
u_1=\mathbf{F}_{des} \cdot \mathbf{z}_B
$$
To determine the other three inputs, we must consider the rotation errors. First, we observe that the desired $z_B$ direction is along the desired thrust vector:
$$
\mathbf{z}_{B, {des}}=\frac{\mathbf{F}_{des}}{\left\|\mathbf{F}_{des}\right\|} .
$$

Thus if $\mathbf{e}_3=[0,0,1]^T$, the desired rotation ${ }^W R_B$ denoted by $R_{d e s}$ for brevity is given by:
$$
R_{des} \mathbf{e}_3=\mathbf{z}_{B, {des}} .
$$

Knowing the specified yaw angle along the trajectory, $\psi_T(t)$, we compute $\mathbf{x}_{B {,des}}$ and $\mathbf{y}_{B {,des}}$ as in the previous section:
$$
\mathbf{x}_{C, d e s}=\left[\cos \psi_T, \sin \psi_T, 0\right]^T,
$$
and
$$
\mathbf{y}_{B, {des}}=\frac{\mathbf{z}_{B, {des}} \times \mathbf{x}_{C, {des}}}{\left\|\mathbf{z}_{B, {des}} \times \mathbf{x}_{C, {des}}\right\|}, \mathbf{x}_{B, {des}}=\mathbf{y}_{B, {des}} \times \mathbf{z}_{B, {des}},
$$
provided $\mathbf{x}_{C {,des}} \times \mathbf{z}_{B {,des}} \neq 0$. This defines the desired rotation matrix $R_{des}$. While mathematically this singularity is a single point in $S O(3)$, this computation results in large changes in the unit vectors in the neighborhood of the singularity. To fix this problem, we observe that $-\mathbf{x}_{B \text {, des }}$ and $-\mathbf{y}_{B {,des}}$ are also consistent with the desired yaw angle and body frame $z$ axis. In practice we simply check which one of the solutions is closer to the actual orientation of the quadrotor in order to calculate the desired orientation, $R_{d e s}$.
Next we define the error on orientation:
$$
\mathbf{e}_R=\frac{1}{2}\left(R_{d e s}^T{ }^W R_B-{ }^W R_B^T R_{d e s}\right)^{\vee}
$$
where $\vee$ represents the vee map which takes elements of $s o(3)$ to $\mathbb{R}^3$. 

The angular velocity error is simply the difference between the actual and desired angular velocity in body frame coordinates:
$$
\mathbf{e}_\omega={ }^B\left[\omega_{\mathcal{B W}}\right]-{ }^B\left[\omega_{\mathcal{B W}, T}\right] .
$$

Now the desired moments and the three remaining inputs are computed as follows:
$$
\left[u_2, u_3, u_4\right]^T=-K_R \mathbf{e}_R-K_\omega \mathbf{e}_\omega,
$$
where $K_R$ and $K_\omega$ are diagonal gain matrices. This allows unique gains to be used for roll, pitch, and yaw angle tracking. 

To summarize, the parameters in the Mellinger controller are the feedback gains, $K_p,K_v,K_R,K_\omega$. 

\textbf{Modification to differentiable version.} In practice, we use the PID to replace the simple linear feedback, so for each feedback gain (take $K_p$ as an example), we have the proportional term gain $K_{p,p}$, integral gain $K_{p,i}$ and difference term gain $K_{p,d}$. For the feedback matrices, we usually use diagonal matrices for simplicity. The first two diagonal elements are for $x,y$ axes, and we put the same gains for the $x, y$ axes. Therefore, take $K_{p,p}$ as example, denote the $\rm {xy}$ axes gain as $K_{p,p}^{\rm xy}$, and $K_{p,p} = \operatorname{diag}(K_{p,p}^{\rm xy},K_{p,p}^{\rm xy},K_{p,p}^{\rm z})$. Therefore, for each set of feedback gain matrices, we have 2 feedback gains for proportional, integral, and difference matrices, respectively. These parameters sums up to $2\times 3\times 4 = 24$ trainable parameters.

\section{Detailed Simulator Setup}
\label{appendix:simu}

\textbf{Initial distribution and simulator action design.} In the simulator, we set the tracking target $p_{{\rm goal}}=[0.0,0.0,1.0]$ and randomly initialize the position $p\sim\mathcal{N}([0.0, 0.0, 1.0], \operatorname{diag}([0.02, 0.02, 0.02]))$. The initial roll, pitch, and yaw angle are randomly sampled from $\mathcal{N}(0.0^\circ, 5.0^\circ)$.
For the simulator action space, we reformulate the action space. The action space is 4-dimensional, including desired vertical thrust $F_z$ and rotational forces $F_r,F_p,F_y$. Then these desired forces are converted to desired propeller forces by the power distribution rules
$$
  \begin{cases}
   F_1 = F_z - F_r/2 + F_p/2+F_y\\
   F_2 = F_z - F_r/2 - F_p/2-F_y\\
   F_3 = F_z + F_r/2 - F_p/2+F_y\\
   F_4 = F_z + F_r/2 + F_p/2-F_y\\
  \end{cases}
  $$
where $F_i$ is the desired force of $i^{\rm th}$ propeller. Compared to the original actions in \texttt{gym-pybullet-drones}, which is the rotor RPM, we change it to the normalized desired motor force. We move the motor force to PWM onboard the crazyflies. The motor force to PWM is extensively studied by previous work, like \cite{shi_neural_2019}.

\end{document}